\def\year{2018}\relax
\documentclass[letterpaper]{article} %DO NOT CHANGE THIS
\usepackage{aaai18}  %Required
\usepackage{times}  %Required
\usepackage{helvet}  %Required
\usepackage{courier}  %Required
\usepackage{url}  %Required
\usepackage{subfig}
\usepackage{multirow}
\usepackage{color}
\usepackage{blindtext}
\usepackage{caption}
\usepackage{graphicx}  %Required
\begin{document}
% The file aaai.sty is the style file for AAAI Press
% proceedings, working notes, and technical reports.
%
\title{Happy Travelers Take Big Pictures: \\A Psychological Study with Machine Learning and Big Data}
\author{Xuefeng Liang, Lixin Fan$^{\dagger}$, Yuen Peng Loh$^{\ddagger}$, Yang Liu, Song Tong\\
IST, Graduate School of Informatics, Kyoto University, 606-8501 Kyoto, Japan\\
$^{\dagger}$ Nokia Technologies, Tampere 33100, Finland \\
$^{\ddagger}$ Centre of Image and Signal Processing, University of Malaya, Kuala Lumpur 50603, Malaysia\\
{\tt\small xliang@i.kyoto-u.ac.jp, lixin.fan@nokia.com } \\ {\tt\small lohyuenpeng@siswa.um.edu.my, liu.yang.24m, tong.song.53w@st.kyoto-u.ac.jp}
}

\twocolumn[{%
\renewcommand\twocolumn[1][]{#1}%
\maketitle
\begin{center}
  \setcaptiontype{figure}% Fake a figure environment
    \subfloat[\label{subfig:temple}]{\includegraphics[height = 0.14\linewidth, width=0.16\linewidth]{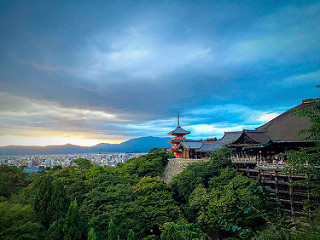}}
	\subfloat[\label{subfig:bridge}]{\includegraphics[height = 0.14\linewidth, width=0.16\linewidth]{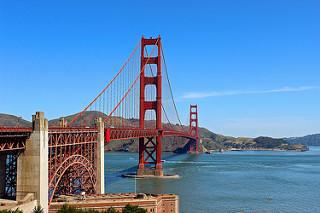}}
	\subfloat[\label{subfig:tower}]{\includegraphics[height = 0.14\linewidth, width=0.16\linewidth]{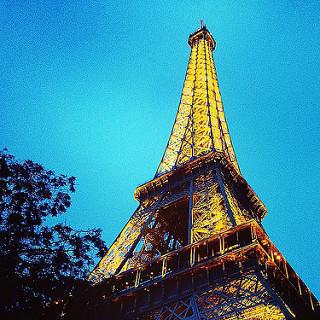}} \hspace{3pt}
	\subfloat[\label{subfig:hollywood}]{\includegraphics[height = 0.14\linewidth, width=0.16\linewidth]{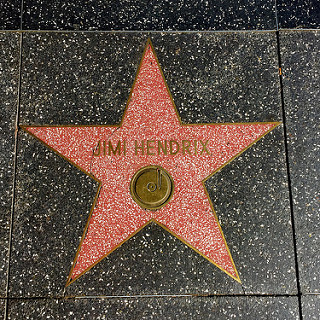}}
	\subfloat[\label{subfig:pis}]{\includegraphics[height = 0.14\linewidth, width=0.16\linewidth]{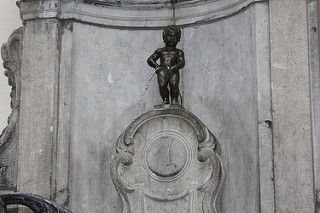}}
	\subfloat[\label{subfig:kimono}]{\includegraphics[height = 0.14\linewidth, width=0.16\linewidth]{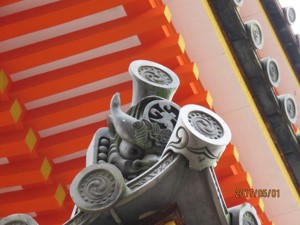}}
    \captionsetup{box=none,parbox=none}
    \captionof{figure}{Wide view angle (a-c) vs. narrow view angle (d-f). The view angle of a travel photo is influenced by the photographer's subconscious emotional state, which is a living example of the psychological broaden-and-build theory.}
\end{center}%
}]

%\twocolumn[{%
%\renewcommand\twocolumn[1][]{#1}%
%\maketitle
%\begin{center}
%    \centering
%    \includegraphics[height = 0.14\linewidth, width=0.16\linewidth]{wide_temple.jpg}\label{subfig:temple}
%	\includegraphics[height = 0.14\linewidth, width=0.16\linewidth]{wide_bridge.jpg}\label{subfig:bridge}
%	\includegraphics[height = 0.14\linewidth, width=0.16\linewidth]{wide_tower.jpg}\label{subfig:tower} \hspace{3pt}
%	\includegraphics[height = 0.14\linewidth, width=0.16\linewidth]{narrow_hollywood.jpg}\label{subfig:hollywood}
%	\includegraphics[height = 0.14\linewidth, width=0.16\linewidth]{narrow_pis.jpg}\label{subfig:pis}
%	\includegraphics[height = 0.14\linewidth, width=0.16\linewidth]{narrow_roof.jpg}\label{subfig:kimono}
%    \captionof{figure}{The view angle of a travel photo encodes the photographer's subconscious emotional state, which is a living example of the psychological broaden-and-build theory.}
%\end{center}
%}]

\begin{abstract}
In psychology, theory-driven researches are usually conducted with extensive laboratory experiments, yet rarely tested or disproved with big data. In this paper, we make use of 418K travel photos with traveler ratings to test the influential ``broaden-and-build'' theory, that suggests positive emotions broaden one's visual attention. The core hypothesis examined in this study is that positive emotion is associated with a wider attention, hence highly-rated sites would trigger wide-angle photographs. By analyzing travel photos, we find a strong correlation between a preference for wide-angle photos and the high rating of tourist sites on TripAdvisor. We are able to carry out this analysis through the use of deep learning algorithms to classify the photos into wide and narrow angles, and present this study as an exemplar of how big data and deep learning can be used to test laboratory findings in the wild.
\end{abstract}

\section{Introduction}
Recent advances in AI technologies (especially Deep Learning) in conjunction with big data offer psychologists an unprecedented opportunity to test theories outside the laboratory. Cognitive scientists and psychologists have been increasingly embracing big data and machine learning to significantly further theory-driven understanding of human behavior and cognition. For example, the sequential dependence functions in higher-order cognition were investigated on millions online reviews posted on Yelp \cite{vinson2016decision}, a machine learning model trained on a standard corpus of online text resulted in human-like semantic biases \cite{caliskan2017semantics}, emerging studies demonstrated that \emph{big data or naturally occurring data sets} (BONDS) can be used as a complement to traditional laboratory paradigms and refine theories \cite{griffiths2015manifesto,goldstone2016discovering,jones2016developing,paxton2017finding}. Following in the footsteps of earlier calls to action, we present here an example of leveraging state-of-the-art machine learning techniques and BONDS as a complement to test psychological theories. Concretely we investigate a real world scenario in which the travelers' photo taking behavior is influenced by a hypothesized psychological mechanism, namely the broaden-and-build theory of positive emotions \cite{fredrickson2004,fredrickson2005positive}.

According to Fredrickson's influential theory, positive emotions broaden (globalize) the attentional scope of the observer and result in processing of a global picture, while negative emotions correlate with a narrowed (localized) attentional focus and induce the processing of local elements. This psychological hypothesis was supported by extensive laboratory experiments \cite{rowe2007positive,tamir2007happy,pourtois2013brain,vanlessen2013positive}. They widely employed a flanker task that required participants to respond to a global-local visual processing task, in which the visual stimuli were either compatible geometric figures / letters or incompatible ones (see supplementary 1 for details). However, to the best of our knowledge, this theory has not been tested with real-world big data. Moreover, it is imprudent to embrace any of these theories blindly since traditional psychological experiments are often conducted in a restricted laboratory environment with limited number of subjects that may result in a considerable bias.

In order to scrutinize the broaden-and-build theory in the travel photo taking scenario, we first develop a deep learning algorithm with a performance in sync with a human, subsequently, cross-check photographers' behaviors by analyzing big data, and address the confounding factors with a set of carefully designed experiments. The results demonstrate that travel photographers' inclination to specific camera viewpoint, e.g. \emph{wide-angle} (figures \ref{subfig:temple}-\ref{subfig:tower}) vs. \emph{narrow-angle} (figures \ref{subfig:hollywood}-\ref{subfig:kimono}) is largely influenced by photographers' emotion at the time of photo taking. Such kinds of influence, which might be subconscious to photographers themselves, nevertheless, is statistically consistent and significant. Roughly speaking, photographers seem to prefer wider-angle photos to narrow-angle ones at high rating tourist sites, while for lower rating sites, the preference appears to be moderate or even going in the reverse direction (see Fig. \ref{fig:siteratio} and Experiments for details). This finding is in accord with the notion of ``positive emotions broaden attention and trigger wide-angle photographs''. Moreover, our study demonstrates a substantial boost of the numbers and diversity of experimental subjects by taking advantages of machine learning techniques and the vast amount of behavior data already available on the internet, which is challenging for traditional laboratory paradigms.

It is our hope that the set up of experiments as well as the proposed deep learning algorithm can be a new method added into the psychologist's toolbox. In addition, the methods adopted in this work have potential significance to real-world applications, such as discovering obscure but high-value tourist sites \cite{zhuang2014anaba}, preventing mental illness of special populations through mining their social media data \cite{stewart2016big} and so on.

%A great deal of our research focus has been placed on the careful experiment design and the development of machine learning algorithms, such that big data collected from photo-sharing websites is unbiased and sound, the proposed algorithm performs the task in sync with a human, the designed experiments carefully test the hypothesis and address the confounding factors.

\section{Materials and Methods}
Here we discuss the data and methods employed to investigate our hypothesis. Specifically, we detail our criteria and procedures in tourist sites selection and photo collection, followed by our proposed machine learning algorithms. %, with evaluations and analysis, which was used for the main experiments.

\begin{figure*}[t] %\vspace{-.1in}
	\centering
	\subfloat[\label{subfig:worldsites}]{\includegraphics[height = 0.2\linewidth, width=0.75\linewidth]{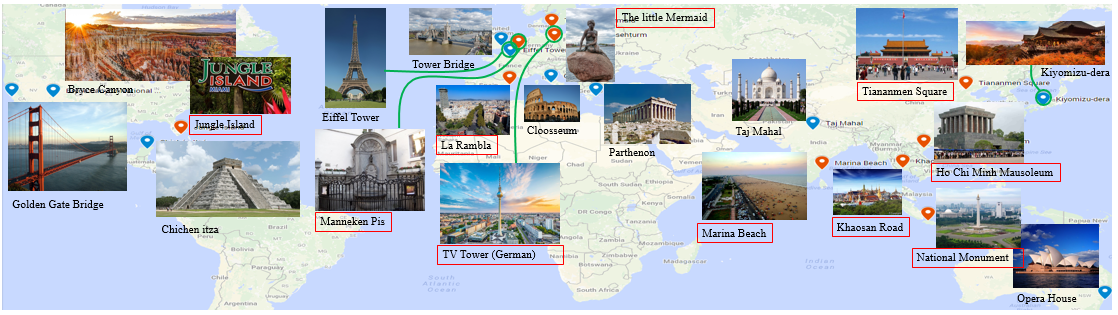}}
    \subfloat[\label{subfig:regionsetes}]{\includegraphics[height = 0.2\linewidth, width=0.25\linewidth]{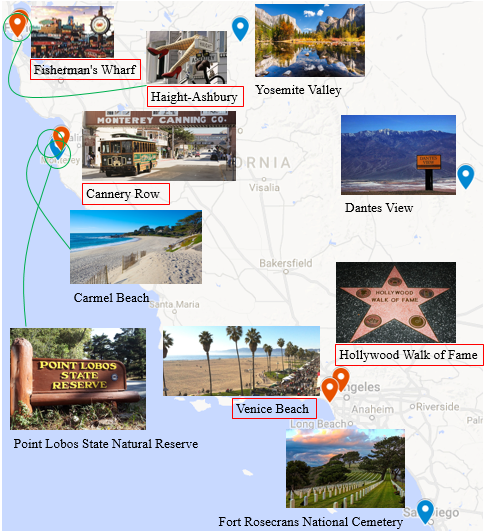}}\vspace{-.1in}
	\caption{(a) The 20 samples of tourist sites across the world, and (b) A zoom-in view of 10 samples in California (USA).}
	\label{fig:SiteLocation} %\vspace{-.1in}
\end{figure*}
\subsection{Tourist Sites Selection}
To test the ``broaden-and-build'' theory using BONDS, we studied the travel photos from selected tourist sites that are hosted on \emph{TripAdvisor} (https://www.tripadvisor.com). The selection is based on five criteria: (1) Popularity: Recommended by top search engines - \emph{TripAdvisor}, \emph{National Geographic} and \emph{Travel + Leisure}; (2) Objectivity: Having at least 1.5K votes for each site regardless of language, age, gender, nationality, etc; (3) Generality: Located across in Asia, Europe, and Americas; (4) Diversity: Keeping site types as diverse as possible, but avoid religious places; (5) Independence: Having an appropriate distance from other sites to avoid cross-rating. Based on their available locations, 70 sites were selected and travel photos associated with these sites were used as our study targets (see supplementary 2 for more details). Figure \ref{fig:ScoreDistribution} illustrates the positions of these 70 sites on the distribution of 12K suitable candidates (the green curve), while Fig. \ref{fig:SiteLocation} shows geo-locations of 30 samples of these sites.

\begin{figure}[h] %\vspace{-.1in}
	\centering
    \includegraphics[width=1\linewidth]{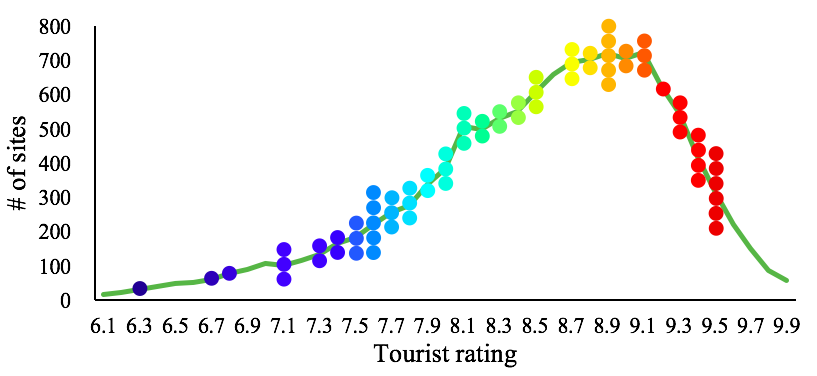} \vspace{-.1in}
	\caption{The 70 selected sites are superimposed on the number distribution of 12K suitable sites w.r.t. site-ratings (the green curve), where the sites with similar ratings are aligned vertically.}
	\label{fig:ScoreDistribution} \vspace{-.1in}
\end{figure}

\subsection{Photo Datasets}
\begin{table}[h]
\scriptsize
    \centering
    %\resizebox{\textwidth}{!}{%
    \caption{Datasets for the estimation of machine learning methods and psychological experiments.} \vspace{-5pt}
    \begin{tabular}{lcccc}
    \hline
    Name & \# of photos & \# of wide-angle & \# of narrow-angle & Source \\ \hline
    $D^1_{train}$ & 43,906 & 25,776 & 18,130 & Flickr \\
    $D^1_{test}$  & 8,722 & 4,141 & 4,581 & TripAdvisor \\
    $D^2$ & 418K & -- & -- & TripAdvisor \\
    $D^3$ & 10,000 & -- & -- & YFCC100m \\ \hline
    \end{tabular} %}
    \label{tab:datasets}\vspace{-.1in}
\end{table}

We used three newly collected datasets in this study, as shown in Table \ref{tab:datasets}. $D^1$ is for the estimation of our proposed machine learning methods, where the training data in $D^1_{train}$ were collected from \textit{Flickr} according to the geo-locations of aforementioned sites. Whereas, the testing data, $D^1_{test}$ were made up of evenly distributed amount of photos from 10 tourist sites, that were randomly collected from \textit{TripAdvisor}. The reason of choosing data from different sources is twofold: (1) to avoid the overlap between the training and testing datasets; (2) photos hosted on \textit{TripAdvisor} were uploaded by travelers who rated the tourist sites, thus the site-ratings and photo contents would be closely related. Due to the second consideration, we created the dataset $D^2$, which consists of 418K travel photos taken at the 70 tourist sites and collected from \textit{TripAdvisor} without overlapping $D^1_{test}$, to test the hypothesized correlation between tourists' positive emotion and the choice of wide-angle photos. The third set, called $D^3$, consists of 10K \textit{random photos} collected from YFCC100m dataset \cite{thomee2016yfcc} without using geo-tag or any other keywords. These photos were used to test the preferences of photo-taking behaviors in a completely random (or neutral emotion) mode.
% ; (3) since we are going to test the hypothesis by making full use of the collected data from \textit{TripAdvisor}, $D^1_{train}$ was made up of \textit{Flickr} photos. Moreover, we noticed that a small portion ($< 8\%$ on average) of \textit{Flickr} data seems not to be the travel photos, which may produce a certain bias in our analysis.

Nevertheless, raw data collected in a completely uncontrolled manner as such are error-prone. The following rectification procedures have been applied to these photos. Firstly, we scrutinized all photos where the erroneously tagged, meaningless, and duplicated photos, i.e. noises, were filtered from the dataset. Secondly, selfies were eliminated from our datasets due to their intrinsic ambiguities, i.e. the attention of such photos are on both the narrow-angle of one or more persons as well as the wide-angle of the background\footnote{The trend of selfies is a relatively recent cultural phenomenon and fast becoming an integral trend in everyday people and also travelers. Though different from the intuition of the current study, it is in our interest to look into it in the future.}.

After rectifying the data, the $D^1$ dataset was then labeled for the estimation of our machine learning algorithms. To build up the training dataset $D^1_{train}$ and testing dataset $D^1_{test}$, about 55K photos were manually labeled as either wide-angle or narrow-angle. We recruited five subjects (5 male, mean age = 28) and designed a binary classification task for them. Before the task, 10 wide-angle and 10 narrow-angle photos were demonstrated to let all subjects have a correct understanding. In the task, 20 photos (4 rows and 5 columns) were simultaneously shown on the screen to give a better visual comparison, and each subject classified photos into two categories. This procedure was iteratively carried out until all photos were checked. After collecting the batch results, we removed those ambiguous photos that had less than 4 consistent votes. By this way, $D^1$ consists of 52,628 photos in total with an almost perfect agreement (Fleiss' kappa $k=0.873$) among five subjects, and was used as the ground truth for the estimation of our proposed methods.

\subsection{Methods}
In order to effectively test our hypothesis on such large dataset $D^2$, we developed two task-optimized machine learning models for wide-angle and narrow-angle classification. This section gives a detailed account of our designs, evaluations, and analysis of said models.
\subsubsection{HVS Model}
\begin{figure}[h]%\vspace{-.25in}
	\centering
	\subfloat[\label{subfig:focus}]{\includegraphics[height = 0.19\linewidth, width=0.25\linewidth]{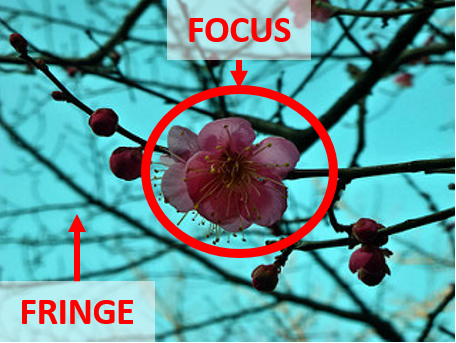}} %\hspace{.05in}
	\subfloat[\label{subfig:scale1}]{\includegraphics[height = 0.19\linewidth, width=0.25\linewidth]{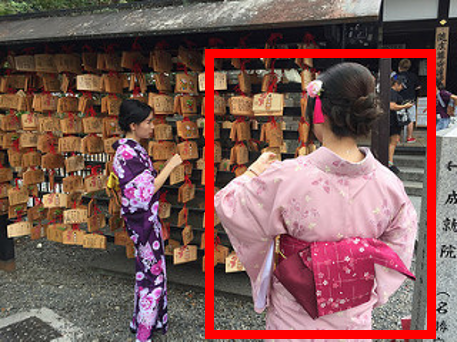}} %\hspace{.1in}
	\subfloat[\label{subfig:scale2}]{\includegraphics[height = 0.19\linewidth, width=0.25\linewidth]{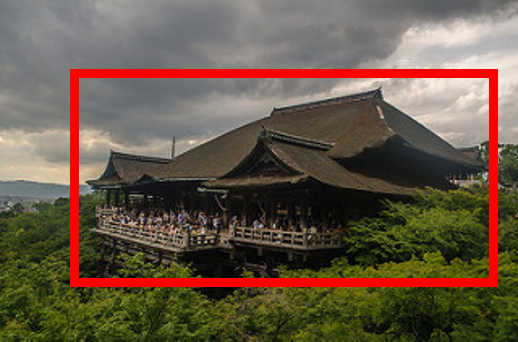}} %\hspace{.1in}
	\subfloat[\label{subfig:scale3}]{\includegraphics[height = 0.19\linewidth, width=0.25\linewidth]{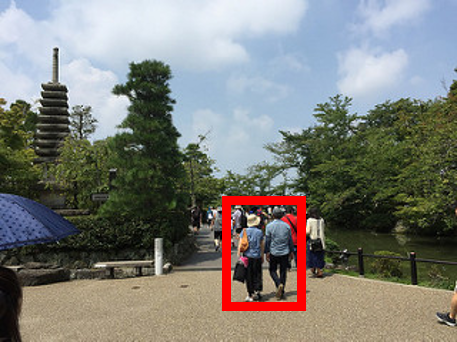}} \vspace{-.1in}
	\caption{Example photos of HVS cues to determine view angle. (a) Narrow-angle of focus lens model, (b) Narrow-angle of spatially large but conceptually small object, (c) Wide-angle of spatially large and conceptually large object, (d) Wide-angle of spatially small and conceptually small object.}
	\label{fig:algoexp}\vspace{-.1in}
\end{figure}

The first model mimics the basics of human visual system (HVS) in determining viewpoints, and is formulated by two cues: a \emph{focus cue} and a \emph{scale cue}. The \emph{focus cue} is based on the finding that a large number of professionally shot close-up view photos adhere to the focus lens model of HVS \cite{tsotsos2011computational} where it focuses on the center object (focus) while the surrounding background is blurred (fringe), as shown in Fig. \ref{subfig:focus}. To model it, we transform images into the frequency domain by using the Non-subsampled Contourlet Transform (NSCT) \cite{da2006nonsubsampled}, in which SURF features \cite{bay2006surf} are extracted and quantized using Fisher Vector \cite{perronnin2010improving}. Afterwards, the classification is implemented by a trained support vector machine (SVM). However, many narrow-angle photos shot by low-cost cameras (e.g. smart phones) do not follow the focus model where entire scene appears sharp, such as Fig. \ref{subfig:scale1}. Therefore, the \emph{scale cue} is derived from observers' ability to differentiate the views by measuring the size of objects, namely the \emph{spatial size} (the object size measured in the photo indicated by the boxes in Fig. \ref{subfig:scale1} and \ref{subfig:scale2} are bigger than the one in Fig. \ref{subfig:scale3}) and the \emph{conceptual size} (the realistic proportion of the object; a person in Fig. \ref{subfig:scale1} is a small object but a building in Fig. \ref{subfig:scale2} is a big object). Referring to Fig. \ref{subfig:scale1} - \ref{subfig:scale3}, a narrow-angle can be determined if the object is spatially large but conceptually small, otherwise, the photo is a wide-angle. We measure the spatial size by an object bounding box proposal method, namely Adobe refined BING boxes \cite{fang2016adobe}. Whereas, the conceptual size is measured by a fine-tuned convolutional neural network (CNN) \cite{krizhevsky2012imagenet}. Hence, this HVS model built following two specific visual cues of human vision can address distinct photo characteristics.

\begin{figure*}[t]
	\centering
	\subfloat{\includegraphics[width=0.7\linewidth]{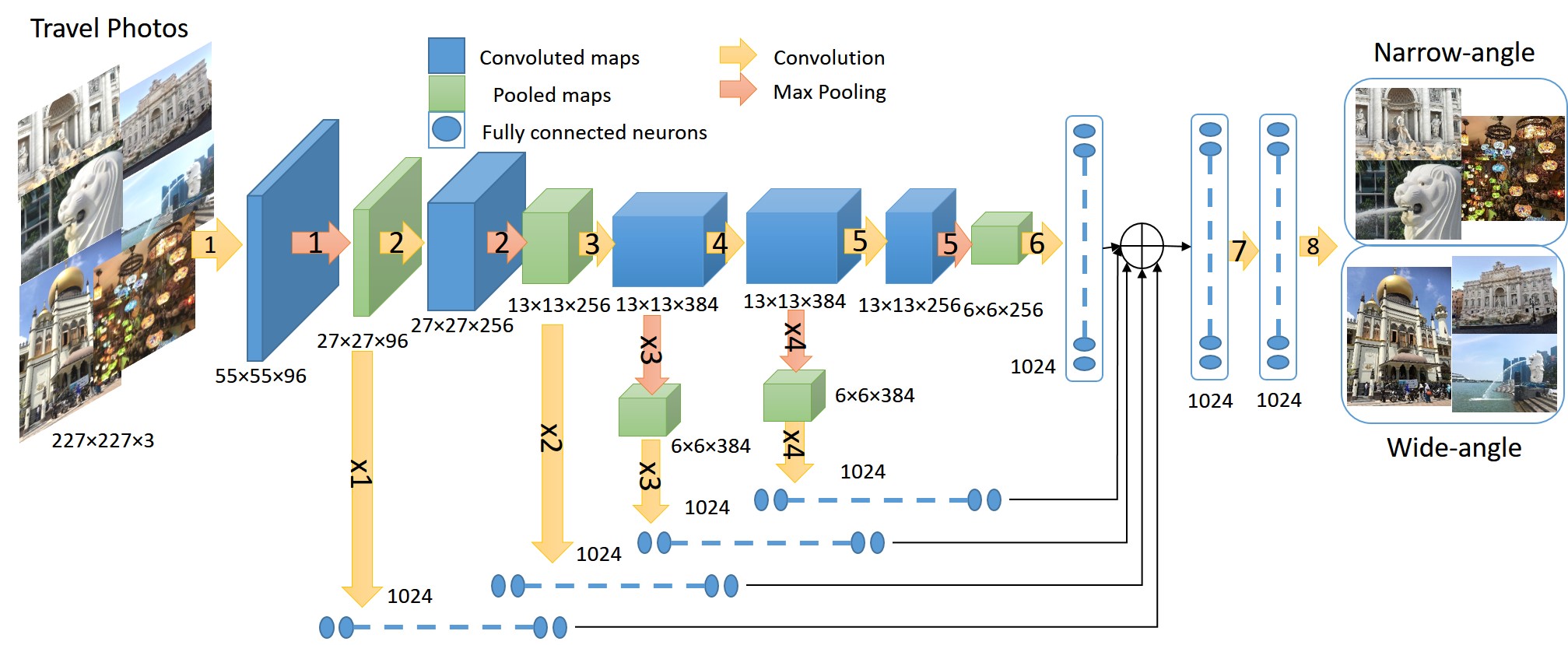}} %\vspace{-.1in} %height = 0.32\linewidth,
	\caption{The architecture of our proposed cumulative feature CNN (CF-CNN). The features from each convolution layer are cumulated into one representation.}
	\label{fig:cnn}\vspace{-.1in}
\end{figure*}

\subsubsection{CF-CNN Model}
We secondly looked into a deep learning technique, by using a single CNN to perform this view angle classification as opposed to the hand-designed HVS model. This is on account of the success shown by CNN at discovering high level features for a variety of tasks \cite{donahue2014decaf,zeiler2014visualizing,yosinski2015understanding,lee2017deep}. However, conventional CNNs only utilize single high level feature after multiple layers of convolution. According to our pilot investigation, features that are crucial for view angle classification may vanish after multiple convolution and pooling operations in the conventional CNNs. Therefore, we designed a cumulative feature CNN (CF-CNN) that extracts features from each stage and accumulates them into one representation, hence incorporating both low and high level features for the classification task. Figure \ref{fig:cnn} illustrates the architecture of our model, where travel photos are the inputs and the outputs are their respective narrow and wide angle categorization. Specifically, we introduced additional convolution paths (convx) on each existing convolution (conv) layer, to produce 1024-dimension features. The new convx layers are placed after pooling layers (pool), and if the conv is not followed by a pool (conv3-conv4), pooling layers are added for them (poolx) before the convx, as shown by the path illustration in Fig. \ref{fig:cnn}. We use the pooling size $\left[3\times 3\right]$ for all pool and poolx. The kernel sizes of conv are transfered from the AlexNet architecture \cite{krizhevsky2012imagenet}, while the convx kernel follows the size of the feature map that is to be convolved, e.g. the feature map after pool1 is $27 \times 27 \times 96$, hence the kernel size of convx1 is $27 \times 27 \times 96$ and mapped to 1024 neurons. These convx layers are to be trained end-to-end with all the other conv layers, thus the kernels are expected to focus on significant features from different levels. Hence, they are directly summed up to obtain the cumulative feature and proceed to the subsequent fully connected layers for classification.

%We secondly looked into a deep learning technique, a single CNN to perform this view angle classification as opposed to the hand-designed HVS model. This is on account of the success shown by CNN at discovering high level features for a variety of tasks \cite{donahue2014decaf,zeiler2014visualizing,yosinski2015understanding,lee2017deep}. Hence, the use of CNN for the full task would be able to overcome the bottle-neck faced by the hand-crafted components. However, instead of specifically designing a new network architecture, we chose to re-train an established and lightweight network \cite{krizhevsky2012imagenet} that is trained on large object data \cite{ILSVRC15}. {\color{blue}This is in consideration of two factors: (1) pre-trained weights as initialization improves the generalization of weights during training for better performance in new tasks \cite{yosinski2014transferable}, and (2) a lightweight network will facilitate future development of portable application.} Figure \ref{fig:cnn} illustrates the CNN architecture, where travel photos are the inputs while the outputs are their respective narrow and wide angle categorization. In order to produce the intended two class results, a new last layer with two outputs is introduced as replacement whose initial weights are randomly drawn from a normal distribution, while the weights of the remaining layers were directly transferred from the pre-trained model.

We used 3/4 of wide and narrow angle photos in $D^1_{train}$ for training while the remaining for validation. During the training process, each training image was augmented by resizing the shorter side to 256 dimensions while maintaining aspect ratio, and then a random cropping and flipping is performed, followed by normalization by subtracting with the average image of the dataset. Finally, a $227 \times 227 \times 3$ dimension image was fed to the network.

This model is trained end-to-end using the stochastic gradient descend approach with training batch size of 230, weight decay of 0.0005, and the learning rate that logarithmically reduces from $\eta = 10^{-5}$ after every training epoch. Additionally, we transferred ImageNet pre-trained weights from the AlexNet for conv1 - conv5 to improve the generalization of the main feature extraction layers of our model \cite{yosinski2014transferable}. In order to prevent over-fitting, the training was stopped at 200 epochs where there was no significant reduction in the trend of the validation error. The difference between the validation and training errors was 0.047, an acceptable range of over-fitting as the validation performance achieved over 80\%. Thus, we proceed to perform later classification experiments using this CF-CNN model.

\subsubsection{Performance Evaluation and Analysis}
The performances of HVS and CF-CNN models were evaluated on dataset $D^1_{test}$, the travel photos of 10 sites collected from \emph{TripAdvisor}. The CF-CNN achieved 88.12\% overall classification accuracy, a major improvement in comparison to the HVS model that only reached 64.06\%. Table \ref{tab:accloc} shows that CF-CNN outperforms HVS model at all sites. Additionally, we show in Table \ref{tab:ratioloc}, the ratio of the wide-angle against narrow-angle photos based on CF-CNN's classification (disregarding accuracy) closely matches the ratio of ground truth. This is an indication that the trained CF-CNN is better than the hand-designed HVS approach with a considerable likeness to a human, therefore, the CF-CNN was used for testing the ``broaden-and-build'' theory with real world data.
%no more than 1 unit difference.

\begin{table}[h]
    \tiny
	\centering
	%\resizebox{\textwidth}{!}{%
	\caption{Classification results of HVS and CNN models on narrow and wide angle photos from the 10 tourist sites.} \vspace{-5pt}
	\begin{tabular}{ccccccc}
		\hline
		\multirow{3}{*}{Site} & \multicolumn{3}{c}{\textbf{HVS model}}&\multicolumn{3}{c}{\textbf{CF-CNN model}} \\ %\cline{2-7}
        & \multicolumn{2}{c}{Correct/Total}& &\multicolumn{2}{c}{Correct/Total}& \\
		&Narrow&Wide& Accuracy&Narrow&Wide&Accuracy \\ \hline
		1 & 74/87 & 237/832 & 37.76\% & 46/87 & 819/832 & 94.12\% \\
		2 & 172/240 & 342/617 & 59.98\% & 129/240 & 590/617 & 83.90\% \\
		3 & 246/305 & 196/518 & 53.71\% & 265/305 & 465/518 & 88.70\% \\
		4 & 255/339 & 233/437 & 62.89\% & 303/339 & 333/437 & 81.96\% \\
		5 & 345/439 & 231/429 & 66.36\% & 353/439 & 396/429 & 86.29\% \\
		6 & 343/405 & 201/425 & 65.54\% & 385/405 & 313/425 & 84.10\% \\
		7 & 420/504 & 145/366 & 64.94\% & 480/504 & 207/366 & 78.97\% \\
		8 & 555/689 & 109/201 & 74.61\% & 654/689 & 176/201 & 93.26\% \\
		9 & 502/648 & 210/269 & 77.64\% & 592/648 & 244/269 & 91.17\% \\
		10 & 717/925 & 18/47 & 75.62\% & 903/925 & 33/47 & 96.30\% \\ \hline
		Total & 3629/4581 & 1958/4141 & 64.06\% & 4110/4581 & 3576/4141 & \textbf{88.12\%} \\ \hline
	\end{tabular} %}
	\label{tab:accloc}%\vspace{-.1in}
\end{table}

\begin{table}[h]
\scriptsize
    \centering
    %\resizebox{\textwidth}{!}{%
    \caption{Comparison of wide-angle and narrow-angle ratio between the ground truth, HVS model, and CF-CNN model.} \vspace{-5pt}
    \begin{tabular}{ccccccccccc}
    \hline
    Site  & 1   & 2   & 3   & 4   & 5   & 6   & 7   & 8   & 9   & 10  \\ \hline
    GT & 9:1 & 7:3 & 6:4 & 6:4 & 5:5 & 5:5 & 4:6 & 2:8 & 3:7 & 1:9 \\
    HVS   & 3:7 & 5:5 & 3:7 & 4:6 & 4:6 & 3:7 & 3:7 & 3:7 & 4:6 & 2:8 \\
    CF-CNN   & 9:1 & 8:2 & 6:4 & 5:5 & 6:4 & 4:6 & 3:7 & 2:8 & 3:7 & 1:9 \\ \hline
    \end{tabular} %}
    \label{tab:ratioloc}%\vspace{-.1in}
\end{table}

%\begin{table}[h]
%	\centering
%	%\resizebox{\textwidth}{!}{%
%	\caption{Comparison of narrow-angle and wide-angle ratio between human, HVS model, and CNN model.} \vspace{-5pt}
%	\begin{tabular}{cccc}
%		\hline
%		Site & Human & HVS & CNN \\ \hline
%		1 & 1:9 & 7:3 & 1:9 \\
%		2 & 3:7 & 5:5 & 2:8 \\
%		3 & 4:6 & 7:3 & 4:6 \\
%		4 & 4:6 & 6:4 & 5:5 \\
%		5 & 5:5 & 6:4 & 4:6 \\
%		6 & 5:5 & 7:3 & 6:4 \\
%		7 & 6:4 & 7:3 & 6:4 \\
%		8 & 8:2 & 7:3 & 7:3 \\
%		9 & 7:3 & 6:4 & 6:4 \\
%		10 & 9:1 & 8:2 & 9:1 \\ \hline
%	\end{tabular} %}
%	\label{tab:ratioloc}
%\end{table}

\begin{figure}[h]
	\centering
	\subfloat{\includegraphics[height = 0.17\linewidth, width=0.24\linewidth]{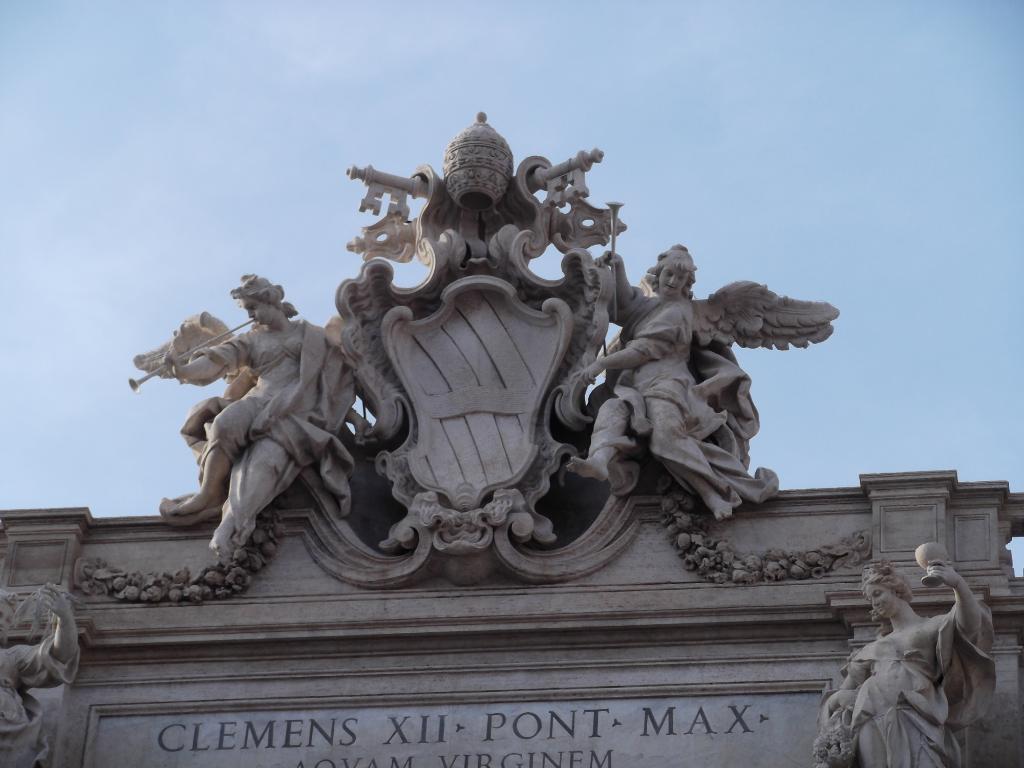}}
	\subfloat{\includegraphics[height = 0.17\linewidth, width=0.24\linewidth]{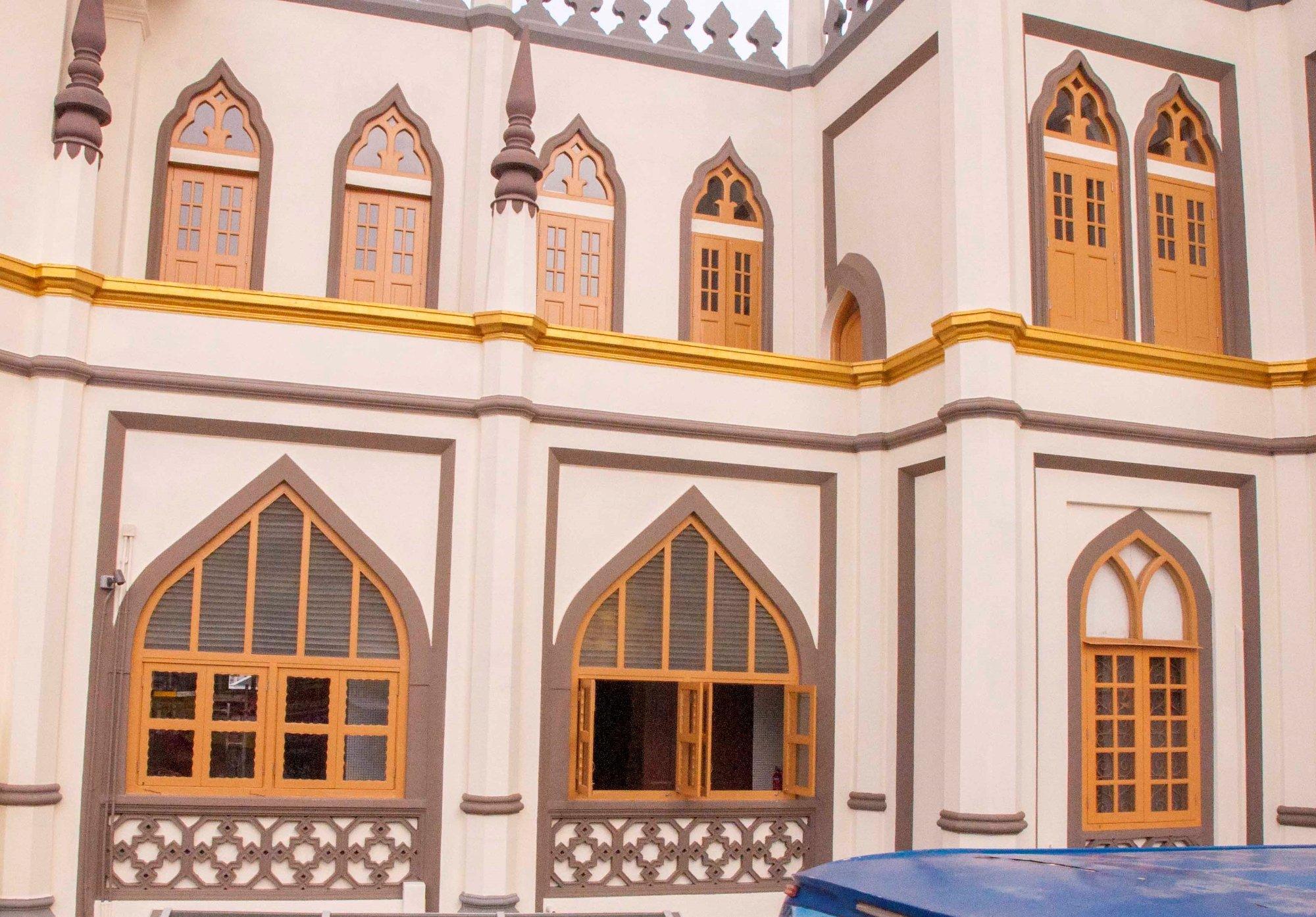}}
	\subfloat{\includegraphics[height = 0.17\linewidth, width=0.24\linewidth]{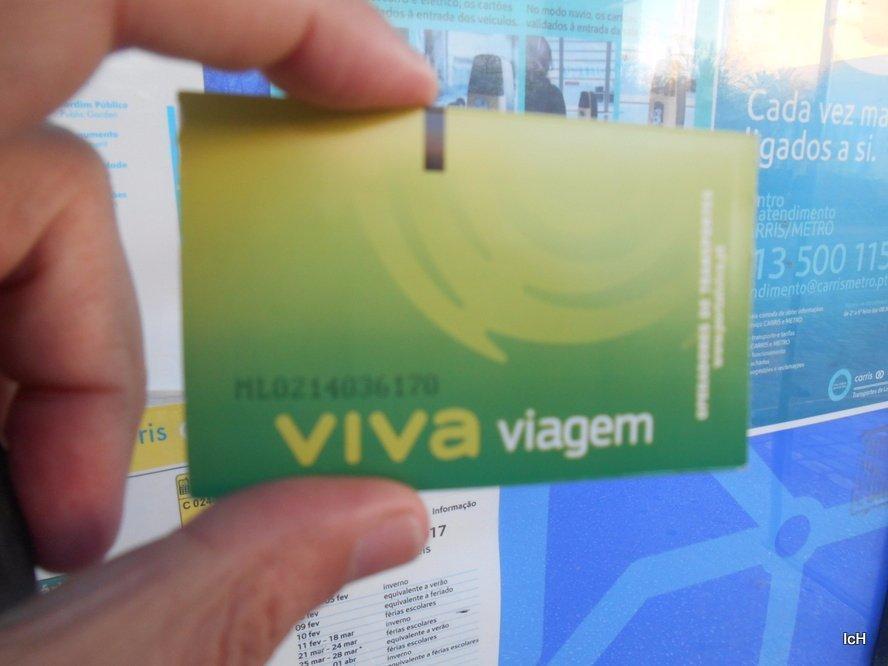}}
	\subfloat{\includegraphics[height = 0.17\linewidth, width=0.24\linewidth]{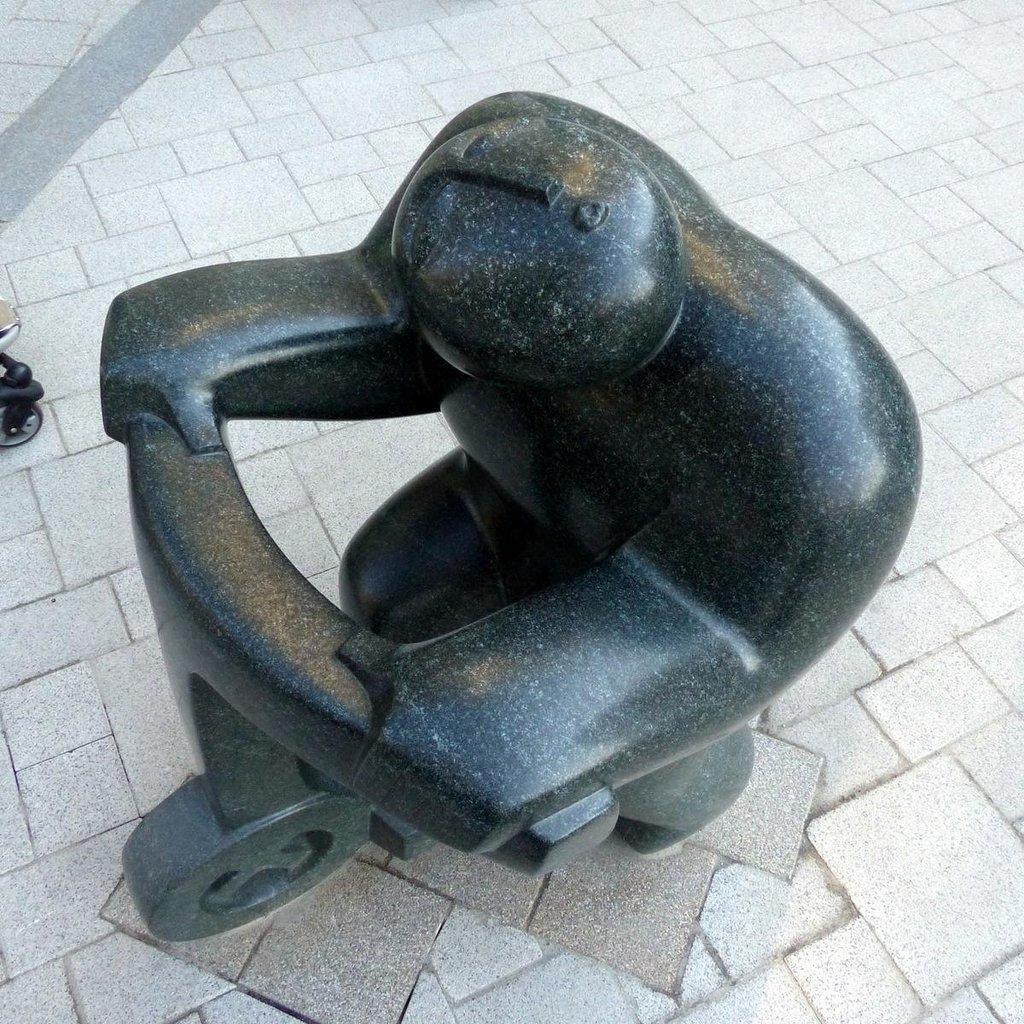}} \\ \vspace{-10pt}
	\subfloat{\includegraphics[height = 0.17\linewidth, width=0.24\linewidth]{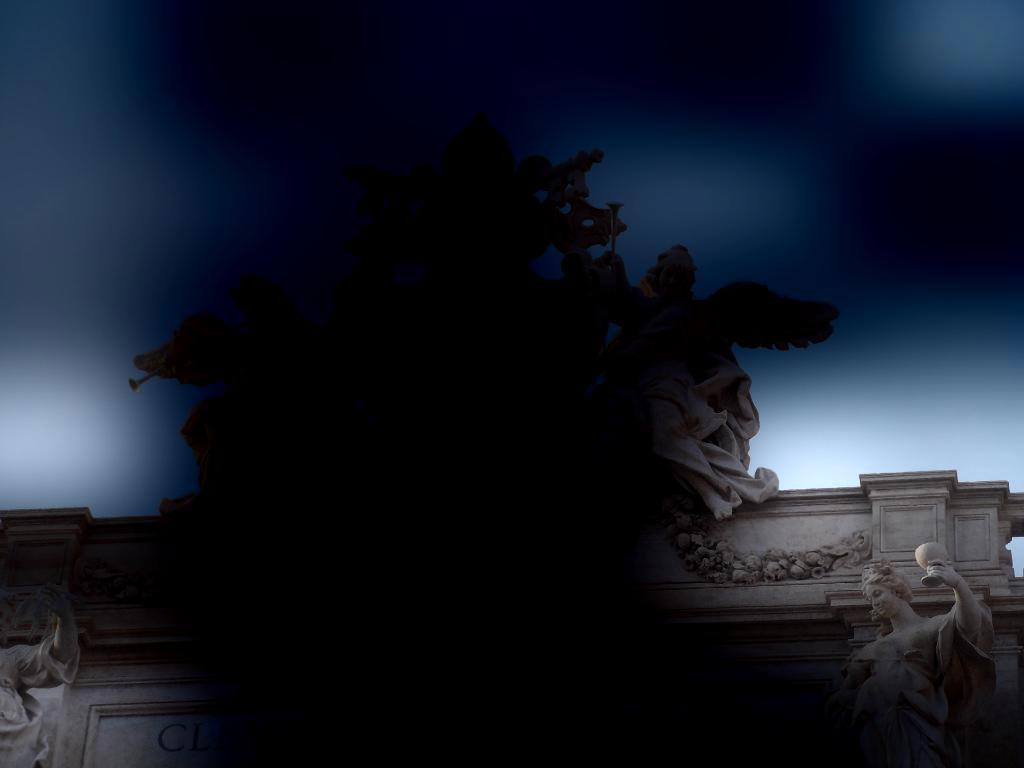}}
	\subfloat{\includegraphics[height = 0.17\linewidth, width=0.24\linewidth]{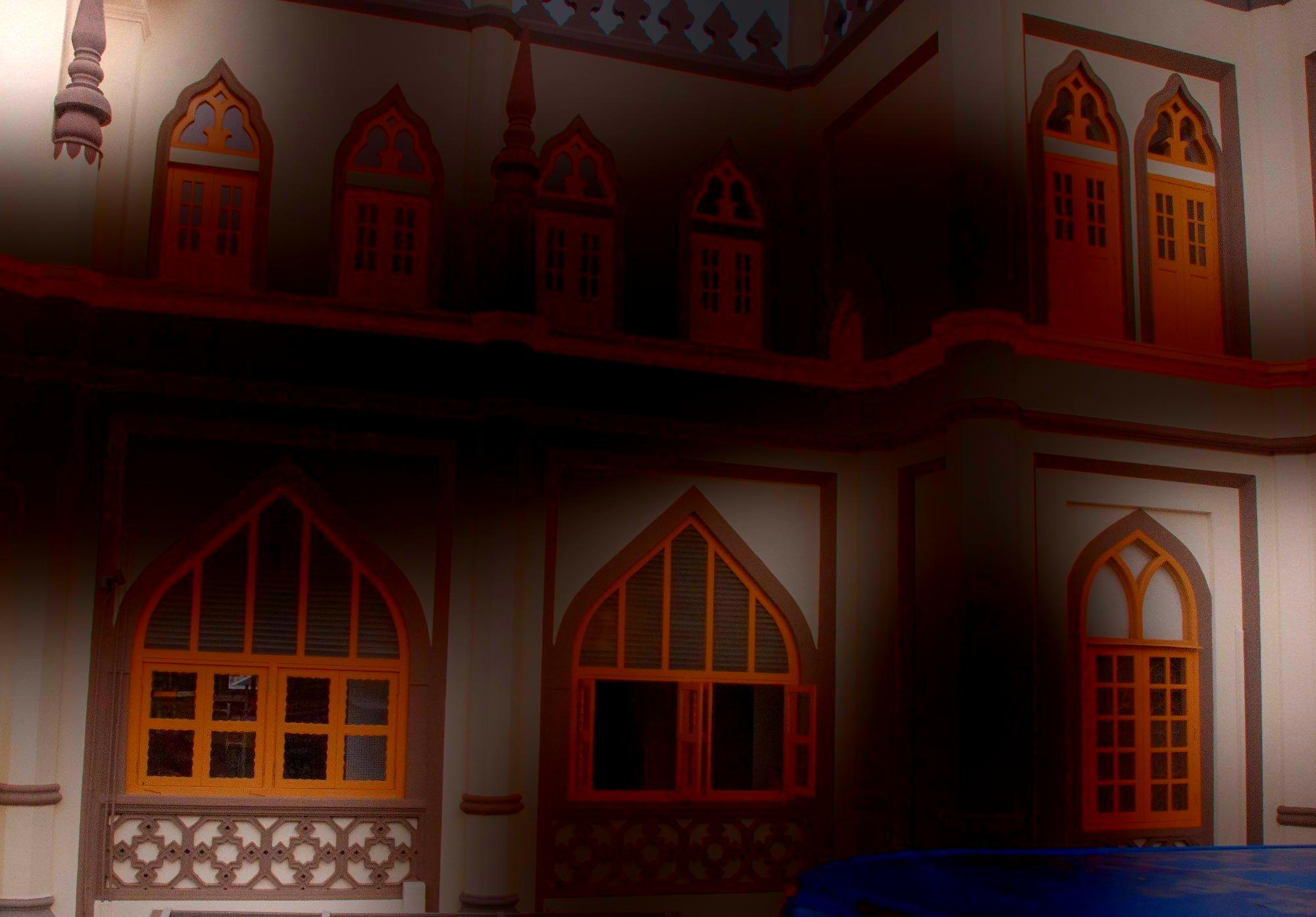}}
	\subfloat{\includegraphics[height = 0.17\linewidth, width=0.24\linewidth]{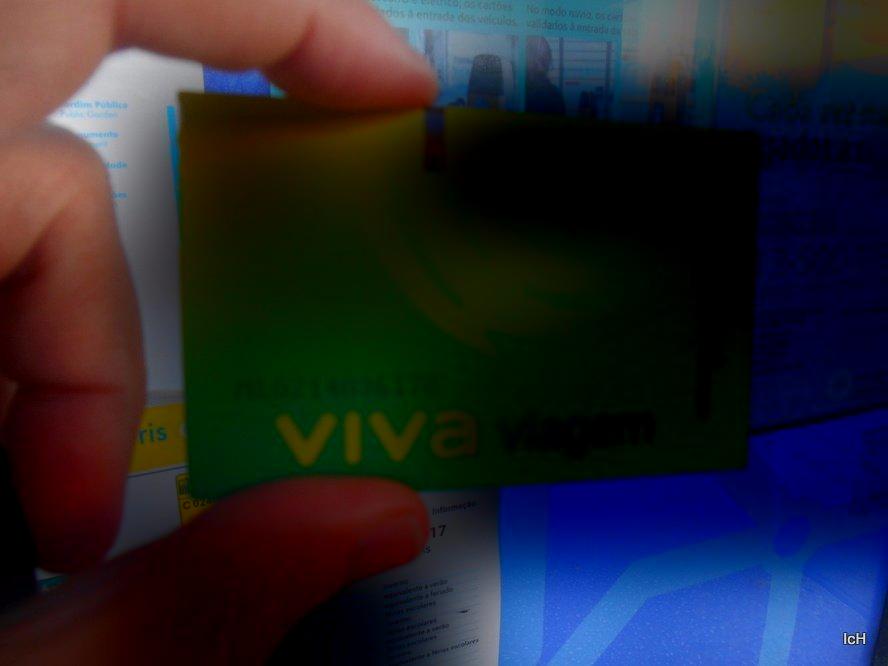}}
	\subfloat{\includegraphics[height = 0.17\linewidth, width=0.24\linewidth]{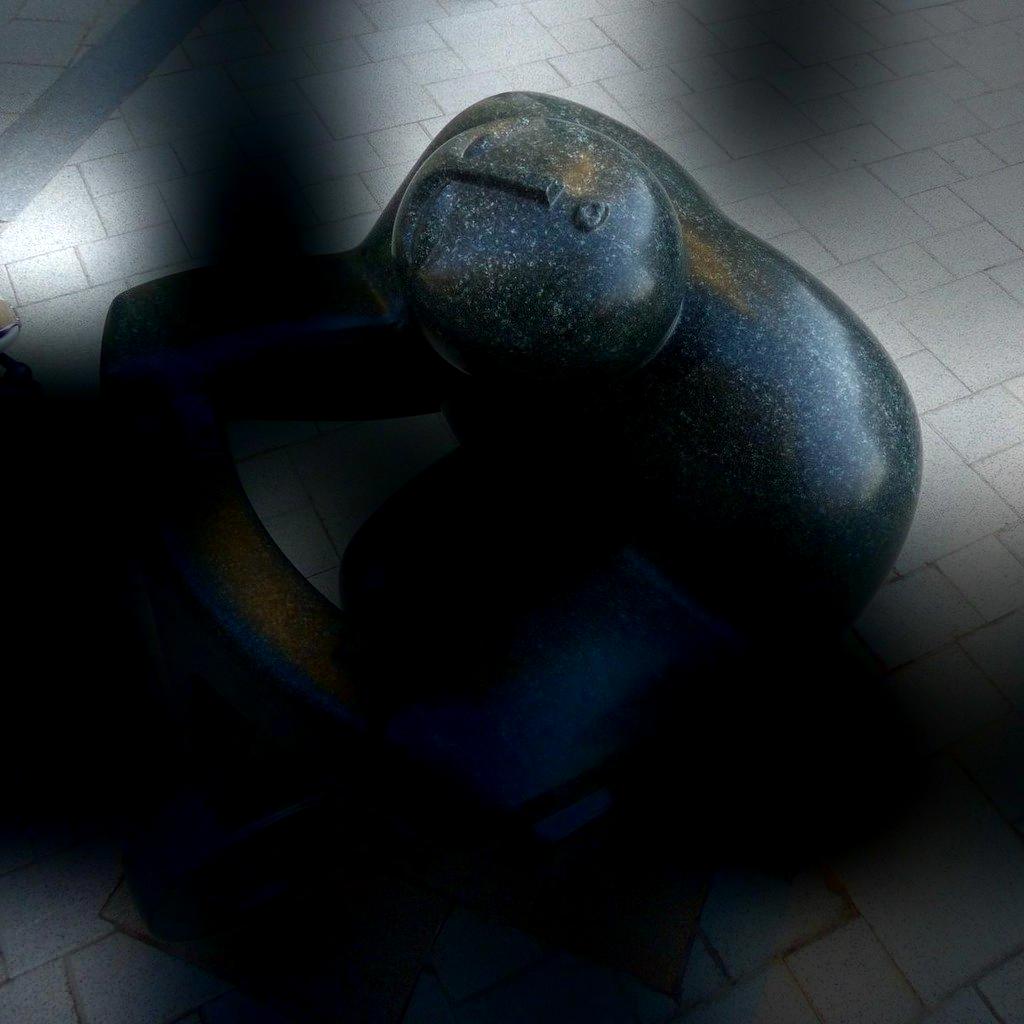}} \\
	
	\subfloat{\includegraphics[height = 0.17\linewidth, width=0.24\linewidth]{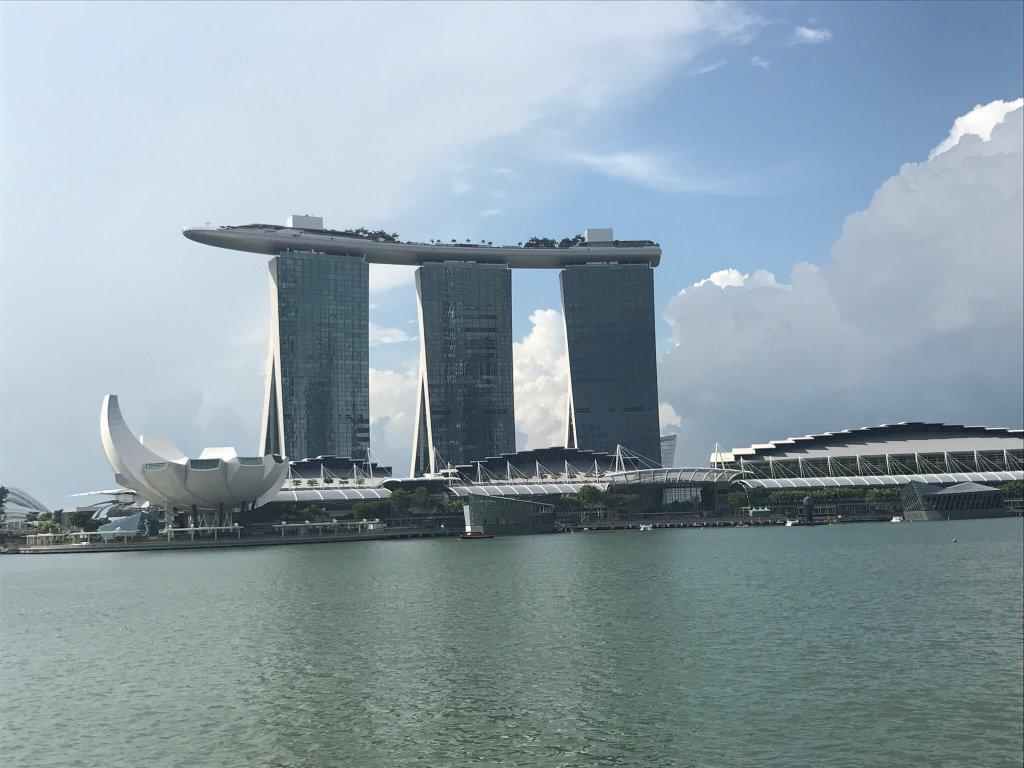}}
	\subfloat{\includegraphics[height = 0.17\linewidth, width=0.24\linewidth]{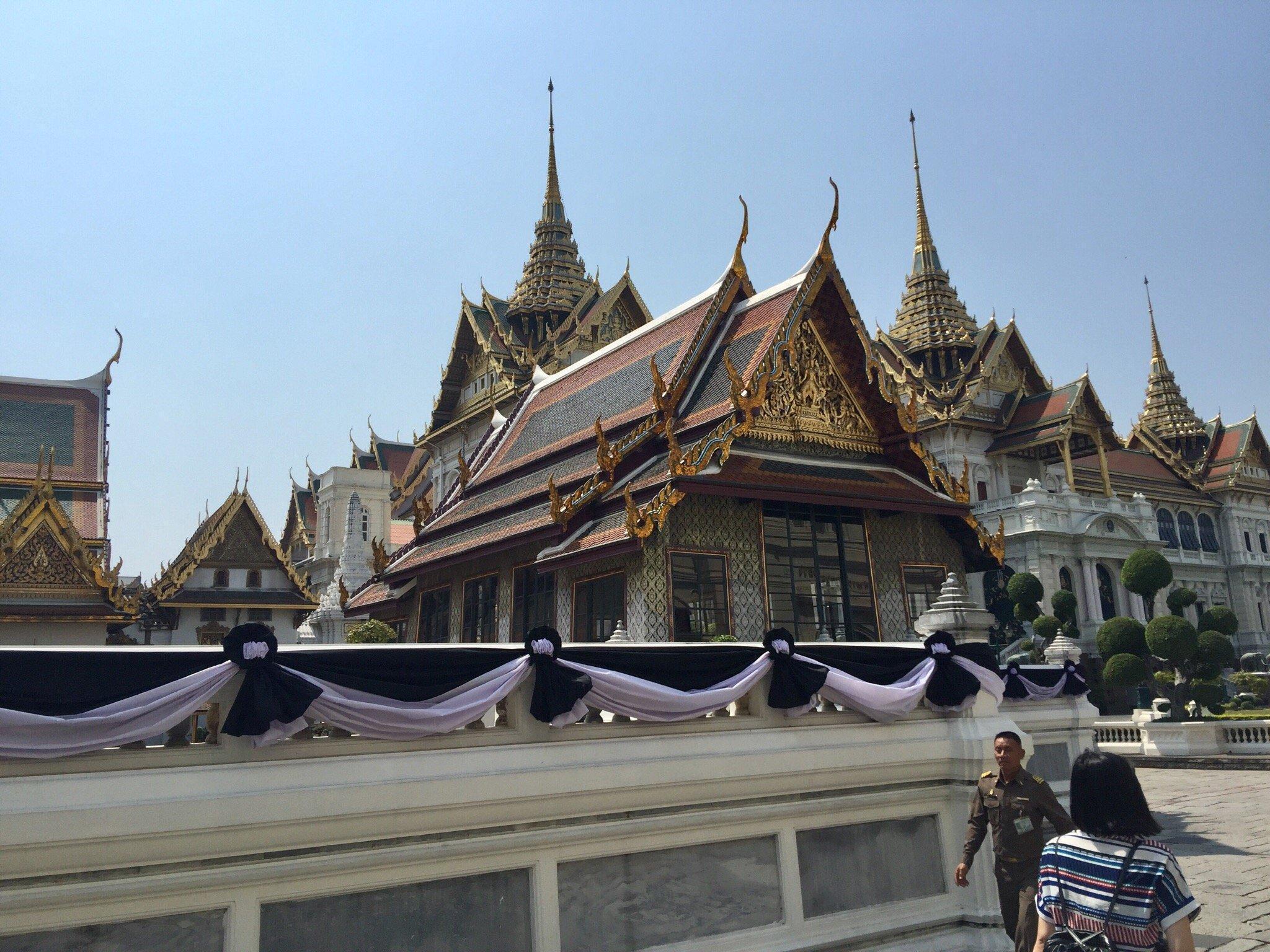}}
	\subfloat{\includegraphics[height = 0.17\linewidth, width=0.24\linewidth]{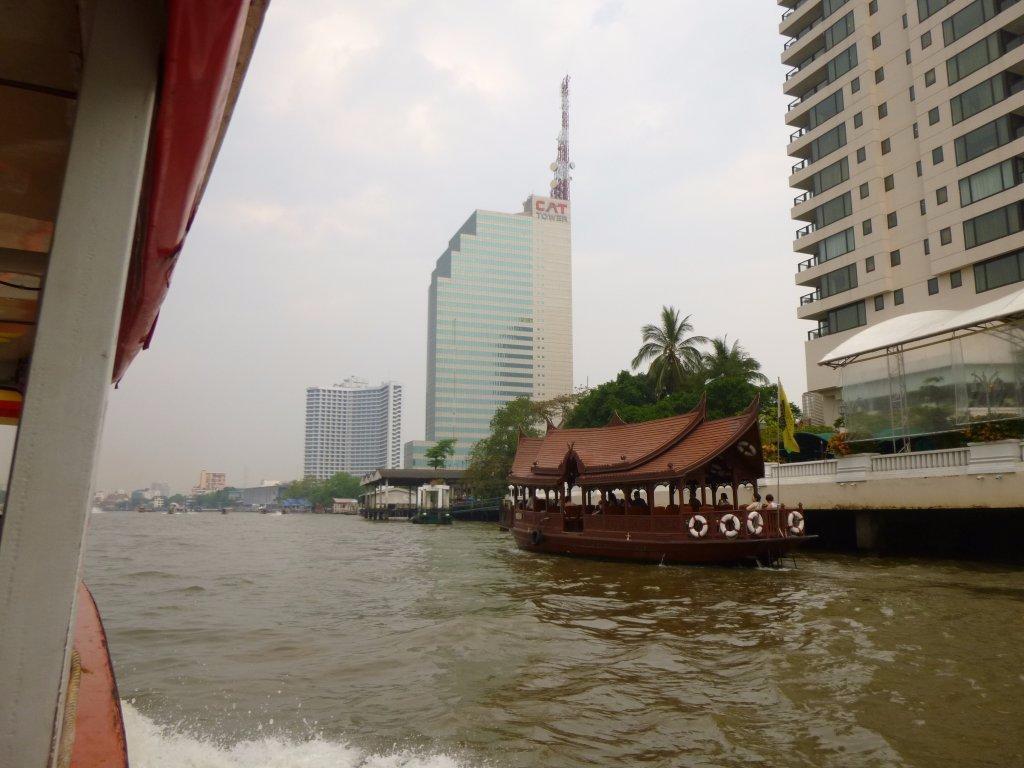}}
	\subfloat{\includegraphics[height = 0.17\linewidth, width=0.24\linewidth]{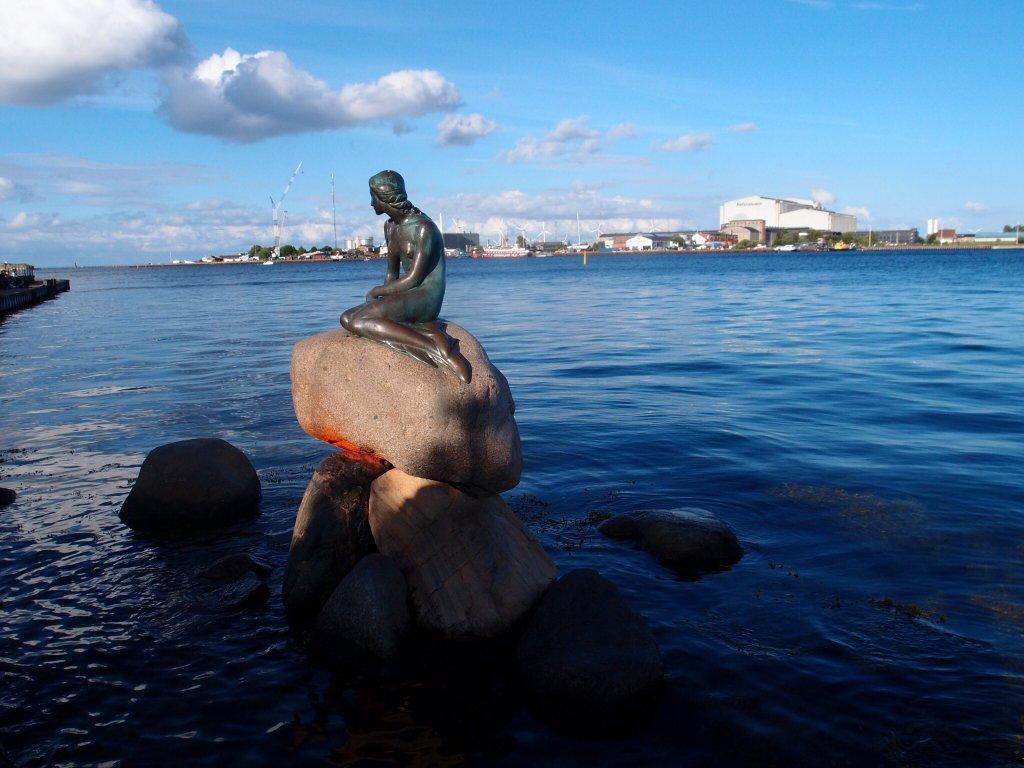}} \\ \vspace{-10pt}
	\subfloat{\includegraphics[height = 0.17\linewidth, width=0.24\linewidth]{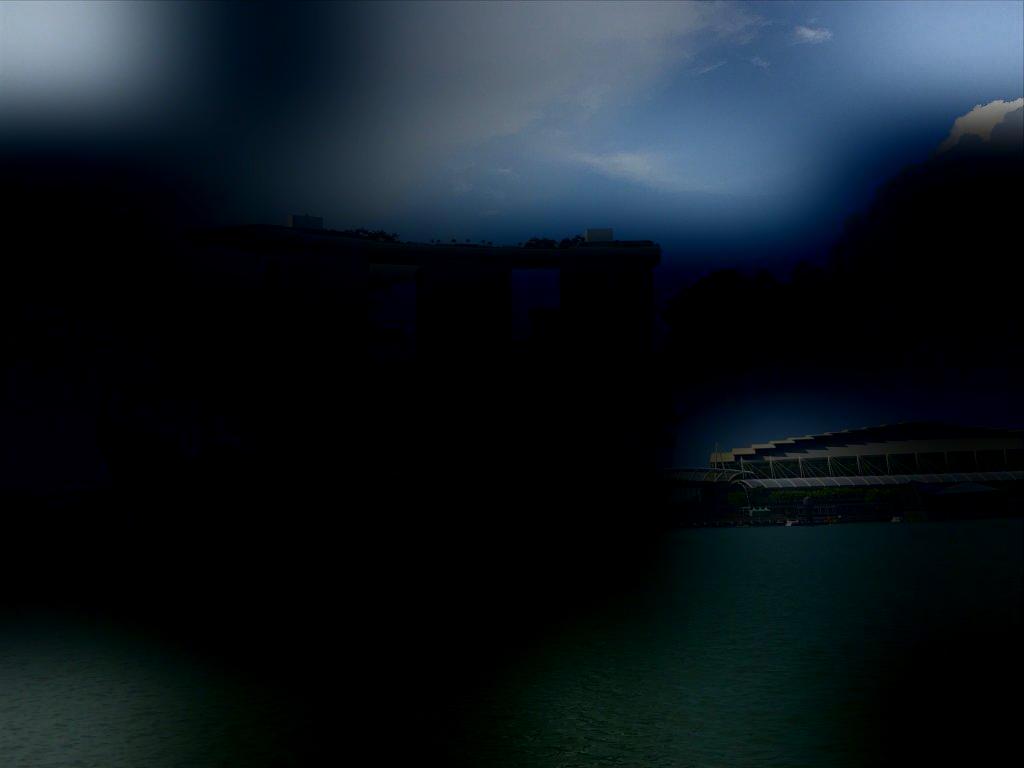}}
	\subfloat{\includegraphics[height = 0.17\linewidth, width=0.24\linewidth]{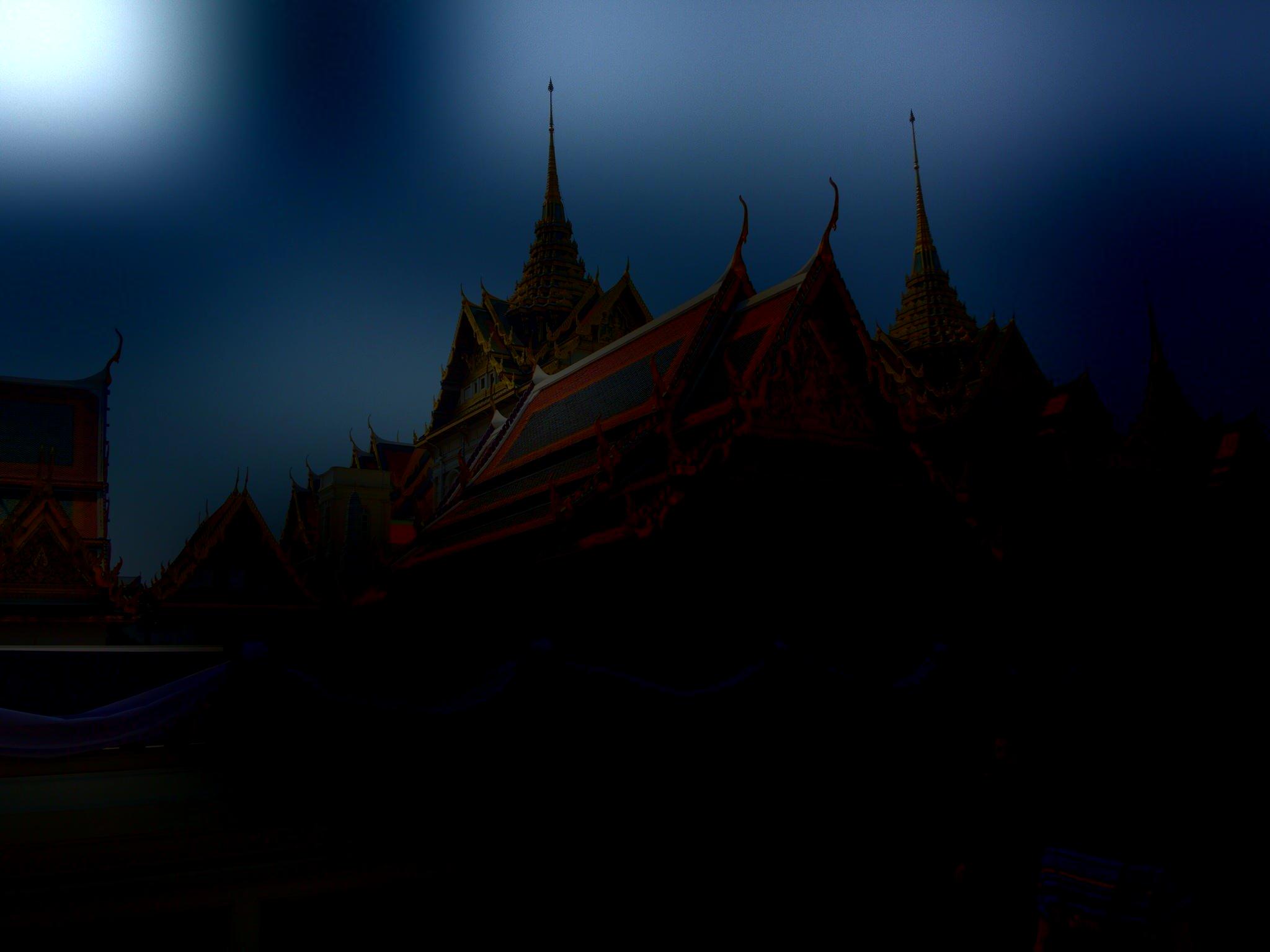}}
	\subfloat{\includegraphics[height = 0.17\linewidth, width=0.24\linewidth]{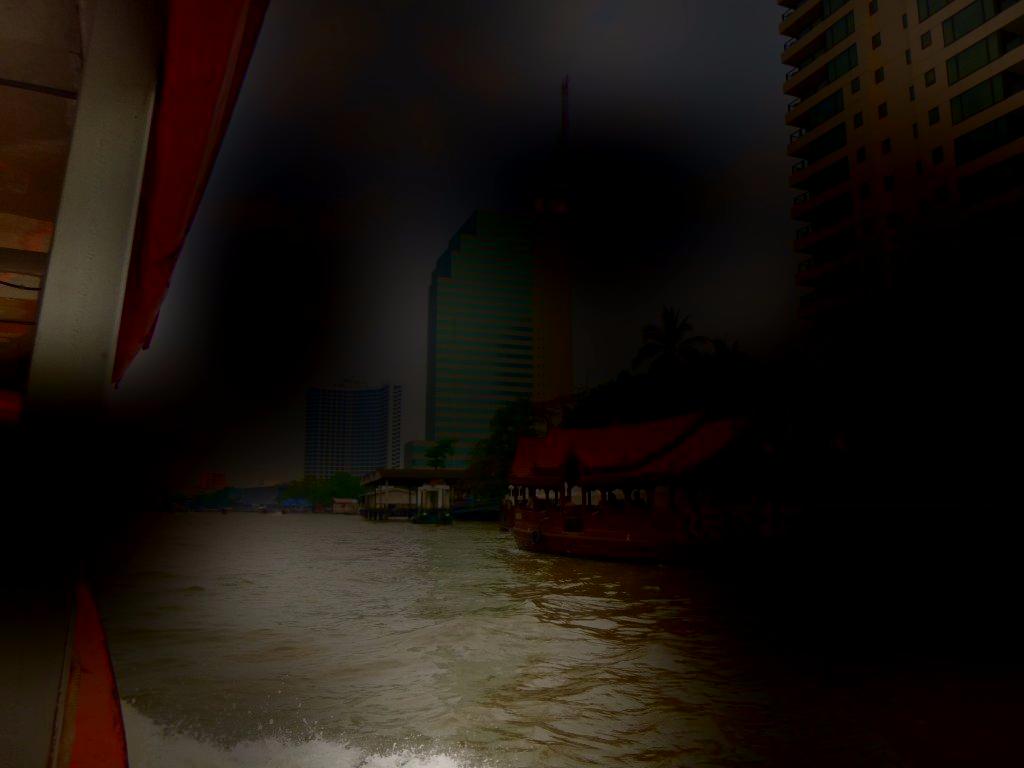}}
	\subfloat{\includegraphics[height = 0.17\linewidth, width=0.24\linewidth]{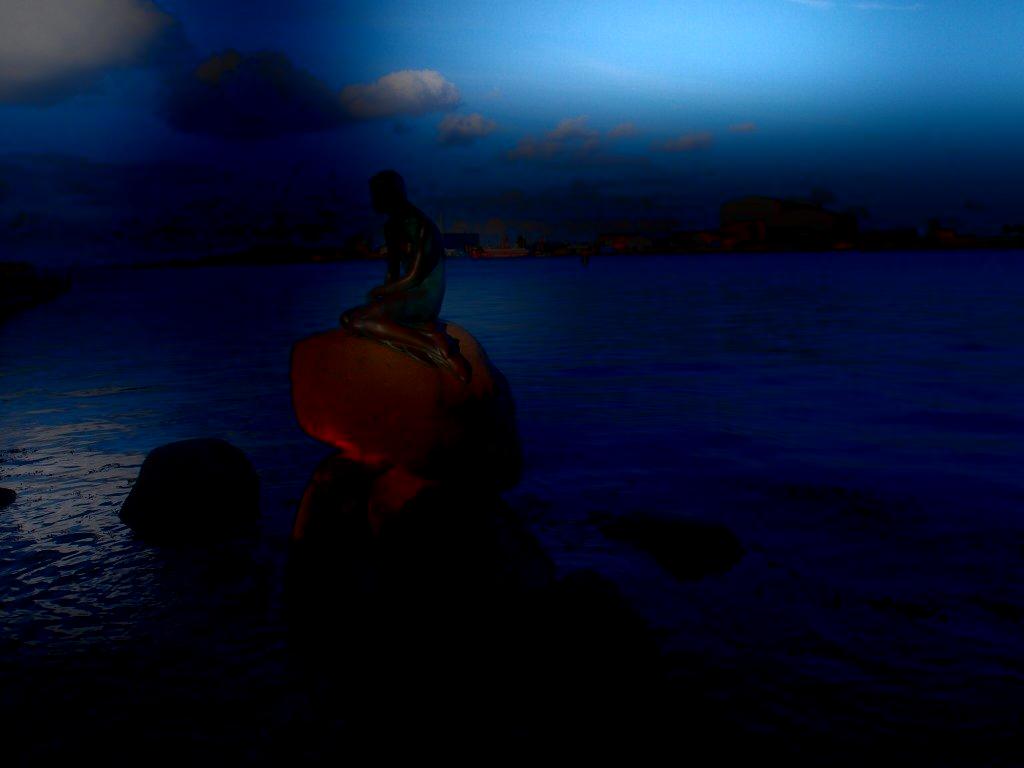}} \\%\vspace{-.1in}
	\caption{Examples of narrow (top) and wide (bottom) angle photos with their high activation regions (bright areas).}
	\label{fig:actmapside}%\vspace{-.1in}
\end{figure}

We also take a further look into the CF-CNN to explore the contributing factors to its performance, by visualizing the last activation maps (highest level features) of the network to find out the spatial location in the photos that is responsible for the classification. The features from lower level layers through convx are not visualized as it has been known that they are less abstract features like edges and high frequency details. Specifically, we extract the activation maps produced by the last pooling operation (pool5) for each test image, where the dimension of the maps are $6 \times 6 \times 256$ as shown in Fig. \ref{fig:cnn}. Max pooling is again performed on the extracted maps in the third dimension to obtain an aggregated map with $6 \times 6$ dimension, where it is then resized to the size of the original image. This final map is used to mask the luminance channel of the original image to obtain a visualization of the area in which the features are used for classification.

This operation has given us an interesting insight of the wide and narrow view angle classification task. Mainly, the border of an image is a major contributor to the classification as opposed to objects only, as we had initially thought. Figure \ref{fig:actmapside} shows several examples where the activations are on the fringe of the image, even though the objects within the image are clearly shown irrespective of the viewpoints. This interesting finding suggests that the strong classification can be achieved by ``looking'' at the image fringe instead of the objects, which goes beyond the focus cue and scale cue designed in the HVS model. We believe that this is one of the component missing from the HVS model that caused its under-performance.

\section{Experiments}\label{sec:experiments}
We hypothesize that prominent tourist sites induces positive emotions to travelers, and subsequently prompt them to capture more wide-angle photos than narrow-angle ones. To test the broaden-and-build theory in photo-taking behaviors, we %designed three experiments to cross-check photographers' behaviors.
structure the analysis to lay out a simple \textit{linear regression} model as follows:
\begin{equation}\label{eq:WRS_model}
{W} = \beta_r {R} + \beta_s {S} + \beta_0,
\end{equation}
in which $W$ is the proportion of wide-angle photos, $R$ and $S$ are the rating score and the (approximated) size of the tourist site, and $\beta_r, \beta_s$ are respective parameters to be estimated, and $\beta_0$ is the offset.

The model (\ref{eq:WRS_model}) is derived from the ``broaden-and-build'' theory based on two assumptions: (1) emotions in the experiments reported here are considered to be represented by traveler ratings on \emph{TripAdvisor}; (2) the scope of attention is naturally or unconsciously manifested by the choice of the view-angle of tourist-taken photos. Note that a competing factor, the site-size $S$, is included in the model because it might also affect the choice of the view-angle. In this paper we adopt the Pearson correlation coefficients (PCC) to quantify and compare the influences of $R$ and $S$ with respect to $W$ (see Tables \ref{tab:exp1_pcc} and \ref{tab:exp2_pcc} below).

First of all, the model (\ref{eq:WRS_model}) is fitted to $W$, $R$ and $S$ of 70 tourist sites, which are elaborated in the supplementary 2. The optimal fitting is reached with parameters ($\beta_r=0.124$, $\beta_s=1.23\mathrm{e}{-4}$, $\beta_0=-0.443$). Note that the relative low R-squared ($\mathrm{R}^2 = 0.646$) of the model indicates a certain amount of data cannot be explained by the model. In order to look for the most influential predictor, we conduct following two experiments.

\subsubsection{Experiment 1: The test with respect to the site-rating} \label{subsec:WorldDataAnalysis}
%The aim of this experiment was to assess how tourist sites dependent emotions would affect the choice of the view angle of travel photos across the world, meanwhile, whether the the local region and culture would be a confounding factor. We did not want to constrain the interpretation of our results with our preconceptions about the broaden-and-build theory. Therefore, 70 tourist sites across the world were selected regardless of religions, races, etc. The associated 418K travel photos in dataset $D^2$ were classified by the CNN model for a statistical analysis. Figure \ref{fig:siteratio} plots the site-rating against the proportion of wide-angle photos at each site.

\begin{figure*}[t] %\vspace{-.2in}
	\centering
    \subfloat[\label{subfig:worldratio}]{\includegraphics[width=0.33\linewidth]{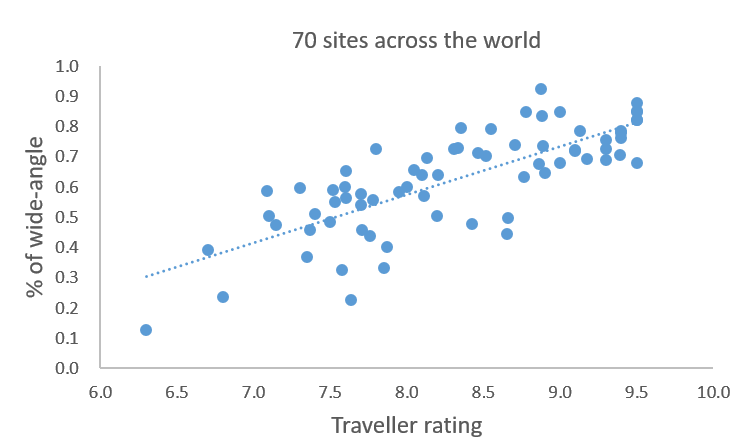}}
    \subfloat[\label{subfig:regionratio}]{\includegraphics[width=0.33\linewidth]{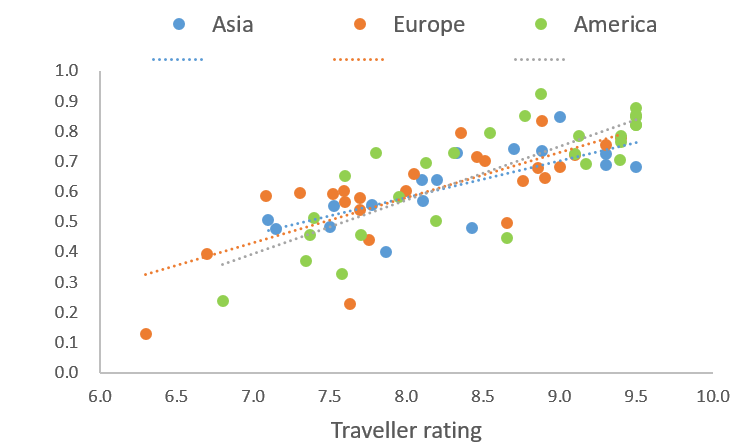}} %\vspace{-.1in} %height = 0.25\linewidth,
    \subfloat[\label{subfig:ratiosize}]{\includegraphics[width=0.33\linewidth]{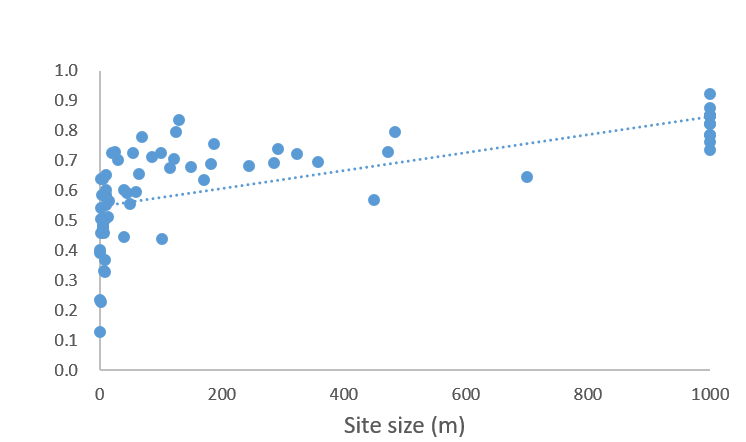}}\vspace{-.1in}
	\caption{(a) A strong correlation between the site-ratings and the proportions of wide-angle photos in the 70 tourist sites from across the world. (b) The analogous correlations appear across Asia, Europe, and Americas. (c) A modest correlation between the site-sizes and the proportions of wide-angle photos.}
	\label{fig:siteratio} \vspace{-.1in}
\end{figure*}

The aim of this experiment is to assess how emotions induced by different tourist sites would affect the choice of the view-angle of travel photos. All 418K travel photos in dataset $D^2$ from 70 tourist sites were classified by the CF-CNN. Figure \ref{fig:siteratio} plots the site-rating against the proportion of wide-angle photos at each site.

\begin{table}[h]
\scriptsize
    \centering
    %\resizebox{\textwidth}{!}{%
    \caption{The Pearson correlation coefficients (PCCs) between the site-rating and the proportion of wide-angle photos with the P-values.} \vspace{-5pt}
    \begin{tabular}{lcccc}
    \hline
      & World-wide & Asia & Europe & Americas \\ \hline
    \# of sites & 70 & 18 & 25 & 27  \\
    PCC   & 0.78 & 0.75 & 0.75 & 0.81 \\
    P-value   & $3.0\mathrm{e}{-15}$ & $5.8\mathrm{e}{-4}$ & $1.9\mathrm{e}{-5}$ & $2.7\mathrm{e}{-7}$ \\ \hline
    \end{tabular} %}
    \label{tab:exp1_pcc}%\vspace{-.1in}
\end{table}

\textbf{Results}: Figure \ref{subfig:worldratio} shows a notable correlation between the site-ratings and the proportions of wide-angle photos across the world. The Pearson correlation coefficient (PCC $= 0.78$ with $P = 3.0\mathrm{e}{-15}$) indicates a strong site-ratings dependent preference. Thus, we deem the site-rating $R$ the principal predictor of the model (\ref{eq:WRS_model}). Not surprisingly, this observation is consistent with the broaden-and-build theory.

To further investigate the influence of local region and culture, these 70 sites are classified into three subgroups according to their geo-locations, Asia, Europe, and Americas. As shown in Table \ref{tab:exp1_pcc} and Fig. \ref{subfig:regionratio}, both the trends of the proportion of wide-angle photos and PCCs of three subgroups are very similar to those of the joint group. Conceivably, the influence of local regions and cultures is negligible.
%No obvious difference is found by comparing to the world-wide sites.

\subsubsection{Experiment 2: The test with respect to the site-size} \label{subsec:LargeSmallAnalysis}
%\subsubsection{Experiment 2: Travel photos from large sites and small sites} \label{subsec:LargeSmallAnalysis}

The choice of view angle of travel photos may also be affected by the size of the tourist site, because people are naturally inclined to take wide-angle photos at a location which has an open space or large object of interest and vice versa. Hence, the size of the site could be a confounding factor as shown in the model (\ref{eq:WRS_model}). The aim of this experiment is to assess the relation between the preferences of photo-taking behaviors and site-sizes. To this end, we define the site-size according to the size of the object/region of interest at the location. For the sites that have the obvious object of interest, we refer to their physical sizes in meters (e.g. statues and buildings). If no such object available, we estimate the size of the region of interest in meters. For the sites with extremely open space (e.g. mountains, canyons, and seashores), their sizes are capped to 1km (see supplementary 2 for details).
%The aim of this experiment is to assess the relation among the preferences of photo-taking behaviors, site-sizes, and traveler emotions.

\begin{table}[h]
\scriptsize
    \centering
    %\resizebox{\textwidth}{!}{%
    \caption{The Pearson correlation coefficients (PCCs) between the site-size and the proportion of wide-angle photos with the P-values.} \vspace{-5pt}
    \begin{tabular}{lcccc}
    \hline
      & All sites & Small sites & Medium sites & Large sites\\ \hline
    \# of sites & 70 & 25 & 14 & 31 \\
    PCC   & 0.62 & 0.44 & 0.33 & 0.61  \\
    P-value   & $6.6\mathrm{e}{-9}$ & 0.026 & 0.024 & 0.002 \\ \hline
    \end{tabular} %}
    \label{tab:exp2_pcc}%\vspace{-.1in}
\end{table}

%\begin{figure*}[t] %\vspace{-.2in}
%	\centering
%%    \subfloat[\label{subfig:largesmallratio}]{\includegraphics[width=0.35\linewidth]{Largesmall.png}}
%    \subfloat[\label{subfig:largeratio}]{\includegraphics[width=0.32\linewidth]{size-ratio.png}} %height = 0.25\linewidth,
%    \subfloat[\label{subfig:smallratio}]{\includegraphics[width=0.32\linewidth]{size-rating.png}}\vspace{-.1in}
%	\caption{The test of correlation with respect to the confounding factor: site size/type. (a) The test of large sites and small sites, where triangles are the mathematical means of two groups. (b) The test of two site types in large group: buildings and seashores. (c) The test of two site types in small group: statues and street views.}
%	\label{fig:largesmall} \vspace{-.1in}
%\end{figure*}

\textbf{Results}: Figure \ref{subfig:ratiosize} and Table \ref{tab:exp2_pcc} illustrate a modest correlation (PCC $= 0.62$) between the size of the site and the proportion of wide-angle photos. This correlation is noticeably weaker, with a margin of 0.16, than the PCC (= 0.78) between the site-rating and the proportion of the wide-angle photos. Since Fig. \ref{subfig:ratiosize} shows that these sites are unevenly distributed according to the site-size, we further look into this factor and separate them into three subgroups, namely \emph{small sites, medium sites}, and \emph{large sites}. Specifically, the sites with size $\leq 10$ meters are in the small group, the sites with size $> 100$ meters are in the large group, and others make up the medium group. We calculate the PCC of each group and list it in Table \ref{tab:exp2_pcc}. These results show even weaker correlations in the three subgroups. In our view, the dwarf influence of the site-size with respect to that of the site-rating reinforces our hypothesis under examination.

Note that the interplay between the site-size, human emotion and photo-taking behavior is twofold: on the one hand, open spaces or large objects make it easy to take wide-angle photos and vice-versa, but are a secondary factor. On the other hand, positive emotions reinforce the tendency for a happy or excited photographer to take wide-angle photos regardless of site-sizes. This ``modulation'' effect is not only in line with the broaden-and-build theory tested in the laboratory, but also suggests that the visual attention is the result of multiple factors.

%Above statistics indicates that both the site-sizes and traveler emotions influence the preference of the view angle of travel photos. On the one hand, the site-size does affect the choice of view angle during photo taking overall. On the other hand, more significant correlation between the site-rating and proportion of wide-angle photos suggests that emotions further modulate the photo-taking behavior above the factor of site-size. This analysis is not only in line with the broaden-and-build theory tested in the laboratory, but also suggests that the visual attention is the result of multiple factors.

\subsubsection{Experiment 3: The test on random photos} \label{subsec:RandomDataAnalysis}
%\subsubsection{Experiment 3: Random photos vs. travel photos from higher and lower rating sites} \label{subsec:RandomDataAnalysis}

While the linear regression model (\ref{eq:WRS_model}) discloses the influences exerted by emotions and site-sizes on photo taking behaviors, the aim of this experiment is to assess the ``default'' behavior in case of a completely random mode (neutral emotion). Therefore, a site independent dataset $D^3$ was randomly collected from YFCC100m dataset \cite{thomee2016yfcc} without using geo-tag or any other keywords, in which 10K photos were classified by the CF-CNN for the statistical analysis. YFCC100m is a subset of Flickr containing 100 million data, which has not only travel photos but a vast diversity (see supplementary 3 for example photos).

\begin{figure}[h] %\vspace{-.1in}
\centering
\includegraphics[width=0.25\textwidth]{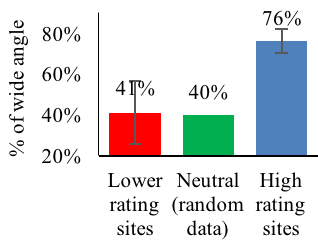}
\caption{The proportions of wide-angle photos in lower rating sites (on average), 10K random data, and high rating sites (on average).}
\label{fig:randonData} \vspace{-.1in}
\end{figure}

In order to have a comparison with other emotional states, we choose the sites, whose ratings are higher than 9.0, as high rating sites according to the site distribution in Fig. \ref{fig:ScoreDistribution} in line with the three-sigma rule in statistics. Whereas, sites with rating lower than 7.4 are termed lower rating sites.

\textbf{Results}: It turns out that, the proportion of the wide-angle photos (the green bar in Fig. \ref{fig:randonData}) in the 10K random data reaches approximately $40\%$. A close investigation of those random photos revealed that a vast majority of narrow-angle photos are clich\'{e} photos of everyday life. The proportion of $40\%$, although slightly in favour of narrow-angle photos, reveals a statistically ``normal behavior'' in composing wide vs. narrow angle photos. Since random photos are, supposedly, taken under neural emotion, this particular ratio serves as a reference and is compared against ratios estimated in other mood states.
%Noticeably the estimated regression model (\ref{eq:WRS_model}) actually predicts the ratio of wide-view photos for random data: by setting $R$ and $S$ to respective average values (7.02, 10.54) of the lower rating sites. The estimated $W (=0.4159)$ is a good approximation of the measured ratio.

For the high rating sites, the average proportion approximates to $76\%$, which is not apparent in lower rating sites. Moreover, the average proportion of $41\%$ for the lower rating sites closely resembles the ratio of random photos. We conjecture that this similarity can be ascribed to the neutral emotion associated with lower rating sites, i.e. such sites are unable to induce positive emotions, subsequently, travelers' photo-taking behaviors are not influenced in a positive manner. On the other hand, the significant high-ratio associated with high rating sites induces photo-taking behavior via broadened visual attention. Another finding is the greater deviation of view proportions in lower rating sites (15.5\%) than high rating ones (6\%). This signifies that the good sites share a consistent ability to induce positive emotions to tourists, which is lacking for lower rating sites.

\section{Conclusion and Discussion}

In this work, we tested the psychological broaden-and-build theory outside the laboratory by leveraging recent machine learning methods and big data from the internet. Our study revealed a strong correlation between the preference for wide-angle photos and the high rating of tourist sites. This preference is ascribed to the notion that positive emotions broaden visual attention and trigger wide-angle photo compositions. Alternatively, neutral emotion induces a slight favor of narrow-angle photos, which is likely associated with those lower rating sites. In addition, by controlling the condition of site-size, our result suggests that the visual attention is the result of multiple factors. We are able to carry out this analysis through the development of a deep learning algorithm for photo view angle classification, which achieves a performance in sync with a human. We hope that the set up of experiments as well as the proposed algorithm can be a new method added into the psychologist's toolbox.

Moreover, the methods adopted in this work have potential significance to real-world applications. For example, recent researches have been focusing on discovering new tourism resources through mining text or evaluating picture quality in SNS. However, few of them tried to link tourists' experiences and mood states with these data, particularly the image data. The broaden-and-build theory with support of real world big data in this study can add a new measure for such task and boost tourism economics. In mental \& welfare heathcare filed, researchers are also reviewing big data resources and their use to characterise applications to address mental illness, e.g. suicide prevention. The other side of the broaden-and-build theory (i.e. negative emotions induce a narrowed attention) with our machine learning method can help such special populations to have better lives through mining their SNS data.

\bibliographystyle{aaai}
\bibliography{PsychologyCollection}

\begin{thebibliography}{}

\bibitem[\protect\citeauthoryear{Bay, Tuytelaars, and
  Van~Gool}{2006}]{bay2006surf}
Bay, H.; Tuytelaars, T.; and Van~Gool, L.
\newblock 2006.
\newblock Surf: Speeded up robust features.
\newblock In {\em European Conference on Computer Vision},  404--417.
\newblock Springer.

\bibitem[\protect\citeauthoryear{Caliskan, Bryson, and
  Narayanan}{2017}]{caliskan2017semantics}
Caliskan, A.; Bryson, J.~J.; and Narayanan, A.
\newblock 2017.
\newblock Semantics derived automatically from language corpora contain
  human-like biases.
\newblock {\em Science} 356(6334):183--186.

\bibitem[\protect\citeauthoryear{Da~Cunha, Zhou, and
  Do}{2006}]{da2006nonsubsampled}
Da~Cunha, A.~L.; Zhou, J.; and Do, M.~N.
\newblock 2006.
\newblock The nonsubsampled contourlet transform: theory, design, and
  applications.
\newblock {\em IEEE Transactions on Image Processing} 15(10):3089--3101.

\bibitem[\protect\citeauthoryear{Donahue \bgroup et al\mbox.\egroup
  }{2014}]{donahue2014decaf}
Donahue, J.; Jia, Y.; Vinyals, O.; Hoffman, J.; Zhang, N.; Tzeng, E.; and
  Darrell, T.
\newblock 2014.
\newblock Decaf: A deep convolutional activation feature for generic visual
  recognition.
\newblock In {\em International Conference on Machine Learning},  647--655.

\bibitem[\protect\citeauthoryear{Fang \bgroup et al\mbox.\egroup
  }{2016}]{fang2016adobe}
Fang, Z.; Cao, Z.; Xiao, Y.; Zhu, L.; and Yuan, J.
\newblock 2016.
\newblock Adobe boxes: Locating object proposals using object adobes.
\newblock {\em IEEE Transactions on Image Processing} 25(9):4116--4128.

\bibitem[\protect\citeauthoryear{Fredrickson and
  Branigan}{2005}]{fredrickson2005positive}
Fredrickson, B.~L., and Branigan, C.
\newblock 2005.
\newblock Positive emotions broaden the scope of attention and thought-action
  repertoires.
\newblock {\em Cognition \& emotion} 19(3):313--332.

\bibitem[\protect\citeauthoryear{Fredrickson}{2004}]{fredrickson2004}
Fredrickson, B.~L.
\newblock 2004.
\newblock The broaden-and-build theory of positive emotions.
\newblock {\em Philosophical Transactions of the Royal Society B: Biological
  Sciences} 359(1449):1367.

\bibitem[\protect\citeauthoryear{Goldstone and
  Lupyan}{2016}]{goldstone2016discovering}
Goldstone, R.~L., and Lupyan, G.
\newblock 2016.
\newblock Discovering psychological principles by mining naturally occurring
  data sets.
\newblock {\em Topics in Cognitive Science} 8(3):548--568.

\bibitem[\protect\citeauthoryear{Griffiths}{2015}]{griffiths2015manifesto}
Griffiths, T.~L.
\newblock 2015.
\newblock Manifesto for a new (computational) cognitive revolution.
\newblock {\em Cognition} 135:21--23.

\bibitem[\protect\citeauthoryear{Jones}{2016}]{jones2016developing}
Jones, M.~N.
\newblock 2016.
\newblock Developing cognitive theory by mining large-scale naturalistic data.
\newblock {\em Big Data in Cognitive Science}  1--12.

\bibitem[\protect\citeauthoryear{Krizhevsky, Sutskever, and
  Hinton}{2012}]{krizhevsky2012imagenet}
Krizhevsky, A.; Sutskever, I.; and Hinton, G.~E.
\newblock 2012.
\newblock Imagenet classification with deep convolutional neural networks.
\newblock In {\em Advances in Neural Information Processing Systems},
  1097--1105.

\bibitem[\protect\citeauthoryear{Lee \bgroup et al\mbox.\egroup
  }{2017}]{lee2017deep}
Lee, S.~H.; Chan, C.~S.; Mayo, S.~J.; and Remagnino, P.
\newblock 2017.
\newblock How deep learning extracts and learns leaf features for plant
  classification.
\newblock {\em Pattern Recognition} 71:1--13.

\bibitem[\protect\citeauthoryear{Paxton and
  Griffiths}{2017}]{paxton2017finding}
Paxton, A., and Griffiths, T.~L.
\newblock 2017.
\newblock Finding the traces of behavioral and cognitive processes in big data
  and naturally occurring datasets.
\newblock {\em Behavior Research Methods}  1--9.

\bibitem[\protect\citeauthoryear{Perronnin, S{\'a}nchez, and
  Mensink}{2010}]{perronnin2010improving}
Perronnin, F.; S{\'a}nchez, J.; and Mensink, T.
\newblock 2010.
\newblock Improving the fisher kernel for large-scale image classification.
\newblock In {\em European Conference on Computer Vision},  143--156.
\newblock Springer.

\bibitem[\protect\citeauthoryear{Pourtois, Schettino, and
  Vuilleumier}{2013}]{pourtois2013brain}
Pourtois, G.; Schettino, A.; and Vuilleumier, P.
\newblock 2013.
\newblock Brain mechanisms for emotional influences on perception and
  attention: what is magic and what is not.
\newblock {\em Biological Psychology}.

\bibitem[\protect\citeauthoryear{Rowe, Hirsh, and
  Anderson}{2007}]{rowe2007positive}
Rowe, G.; Hirsh, J.~B.; and Anderson, A.~K.
\newblock 2007.
\newblock Positive affect increases the breadth of attentional selection.
\newblock {\em Proceedings of the National Academy of Sciences}
  104(1):383--388.

\bibitem[\protect\citeauthoryear{Stewart and Davis}{2016}]{stewart2016big}
Stewart, R., and Davis, K.
\newblock 2016.
\newblock 'big data'in mental health research: current status and emerging
  possibilities.
\newblock {\em Social psychiatry and psychiatric epidemiology}
  51(8):1055--1072.

\bibitem[\protect\citeauthoryear{Tamir and Robinson}{2007}]{tamir2007happy}
Tamir, M., and Robinson, M.~D.
\newblock 2007.
\newblock The happy spotlight: Positive mood and selective attention to
  rewarding information.
\newblock {\em Personality and Social Psychology Bulletin} 33(8):1124--1136.

\bibitem[\protect\citeauthoryear{Thomee \bgroup et al\mbox.\egroup
  }{2016}]{thomee2016yfcc}
Thomee, B.; Shamma, D.~A.; Friedland, G.; Elizalde, B.; Ni, K.; Poland, D.;
  Borth, D.; and Li, L.-J.
\newblock 2016.
\newblock Yfcc100m: The new data in multimedia research.
\newblock {\em Communications of the ACM} 59(2):64--73.

\bibitem[\protect\citeauthoryear{Tsotsos}{2011}]{tsotsos2011computational}
Tsotsos, J.~K.
\newblock 2011.
\newblock {\em A computational perspective on visual attention}.
\newblock MIT Press.

\bibitem[\protect\citeauthoryear{Vanlessen \bgroup et al\mbox.\egroup
  }{2013}]{vanlessen2013positive}
Vanlessen, N.; Rossi, V.; De~Raedt, R.; and Pourtois, G.
\newblock 2013.
\newblock Positive emotion broadens attention focus through decreased
  position-specific spatial encoding in early visual cortex: Evidence from
  erps.
\newblock {\em Cognitive, Affective, \& Behavioral Neuroscience} 13(1):60--79.

\bibitem[\protect\citeauthoryear{Vinson, Dale, and
  Jones}{2016}]{vinson2016decision}
Vinson, D.~W.; Dale, R.; and Jones, M.~N.
\newblock 2016.
\newblock Decision contamination in the wild: Sequential dependencies in yelp
  review ratings.
\newblock In {\em Proceedings of the 38th Annual Meeting of the Cognitive
  Science Society},  1433--1438.

\bibitem[\protect\citeauthoryear{Yosinski \bgroup et al\mbox.\egroup
  }{2014}]{yosinski2014transferable}
Yosinski, J.; Clune, J.; Bengio, Y.; and Lipson, H.
\newblock 2014.
\newblock How transferable are features in deep neural networks?
\newblock In {\em Advances in Neural Information Processing Systems},
  3320--3328.

\bibitem[\protect\citeauthoryear{Yosinski \bgroup et al\mbox.\egroup
  }{2015}]{yosinski2015understanding}
Yosinski, J.; Clune, J.; Nguyen, A.; Fuchs, T.; and Lipson, H.
\newblock 2015.
\newblock Understanding neural networks through deep visualization.
\newblock {\em arXiv preprint arXiv:1506.06579}.

\bibitem[\protect\citeauthoryear{Zeiler and
  Fergus}{2014}]{zeiler2014visualizing}
Zeiler, M.~D., and Fergus, R.
\newblock 2014.
\newblock Visualizing and understanding convolutional networks.
\newblock In {\em European Conference on Computer Vision},  818--833.
\newblock Springer.

\bibitem[\protect\citeauthoryear{Zhuang \bgroup et al\mbox.\egroup
  }{2014}]{zhuang2014anaba}
Zhuang, C.; Ma, Q.; Liang, X.; and Yoshikawa, M.
\newblock 2014.
\newblock Anaba: An obscure sightseeing spots discovering system.
\newblock In {\em International Conference on Multimedia and Expo}.

\end{thebibliography}

\end{document}